\documentclass[10pt,journal,cspaper,compsoc]{IEEEtran}
\usepackage{epsfig}
\usepackage{graphicx}
\usepackage{amsmath}
\usepackage{amssymb}
\usepackage{amsthm}
\usepackage{array}
\usepackage{url}
\usepackage{color}
\usepackage{multirow}
\usepackage{booktabs}

\usepackage{algorithm}
\usepackage{algorithmic}
\usepackage[pagebackref=true,breaklinks=true,colorlinks,citecolor=blue,linkcolor=blue,bookmarks=false]{hyperref}

\newcommand{\etal}{{et al.}}

\newcommand{\ignore}[1]{}
\ifCLASSOPTIONcompsoc
\else
\fi

\ifCLASSINFOpdf
\else
\fi

\begin{document}
%
\title{Physics-Based Generative Adversarial Models for Image Restoration and Beyond}
%
%
%
%

\author{Jinshan~Pan, Jiangxin Dong, Yang Liu, Jiawei Zhang, \\Jimmy Ren, Jinhui Tang, Yu-Wing Tai,
        and~Ming-Hsuan Yang
\IEEEcompsocitemizethanks{\IEEEcompsocthanksitem
J. Pan and J. Tang are with School of Computer Science and Engineering, Nanjing University of Science and Technology, Nanjing, 210094, China.
E-mail: sdluran@gmail.com, jinhuitang@njust.edu.cn.
\IEEEcompsocthanksitem J. Dong and Y. Liu are with School of Mathematical and Sciences, Dalian University of Technology, Dalian, 116024, China.
E-mail: \{lewisyangliu, dongjxjx\}@gmail.com.
\IEEEcompsocthanksitem J. Zhang and J. Ren are with SenseTime Research, Shenzhen, 518000, China. E-mail: zhjw1988@gmail.com, rensijie@sensetime.com.
\IEEEcompsocthanksitem Y.-W. Tai is with the Tencent, Shenzhen, 518054, China. E-mail: yuwingtai@tencent.com.
\IEEEcompsocthanksitem M.-H. Yang is with School of Engineering, University of California, Merced, CA, 95344. E-mail: mhyang@ucmerced.edu.
}
\thanks{}}

\markboth{IEEE Transactions on Pattern Analysis and Machine Intelligence}%
{Pan \MakeLowercase{\textit{et al.}}: Regular Paper}

\IEEEcompsoctitleabstractindextext{%
\begin{abstract}
We present an algorithm to directly solve numerous image restoration problems (e.g., image deblurring, image dehazing, and image deraining).
These problems are ill-posed, and the common assumptions for existing methods are usually based on heuristic image priors.
In this paper, we show that these problems can be solved by generative models with adversarial learning.
However, a straightforward formulation based on a straightforward generative adversarial network (GAN) does not
perform well in these tasks, and some structures of the estimated images are usually not preserved well.
Motivated by an interesting observation that the estimated results should be consistent with the observed inputs under the physics models, we propose an algorithm that guides the estimation process of a specific task within the GAN framework.
The proposed model is trained in an end-to-end fashion and can be applied to a variety of image restoration and low-level vision problems.
Extensive experiments demonstrate that the proposed method performs favorably against state-of-the-art algorithms.
\end{abstract}

}

\maketitle

\IEEEdisplaynotcompsoctitleabstractindextext

%
\IEEEpeerreviewmaketitle

\begin{figure*}[t]
%
\begin{center}
\begin{tabular}{cccccc}
\includegraphics[width = 0.32\linewidth]{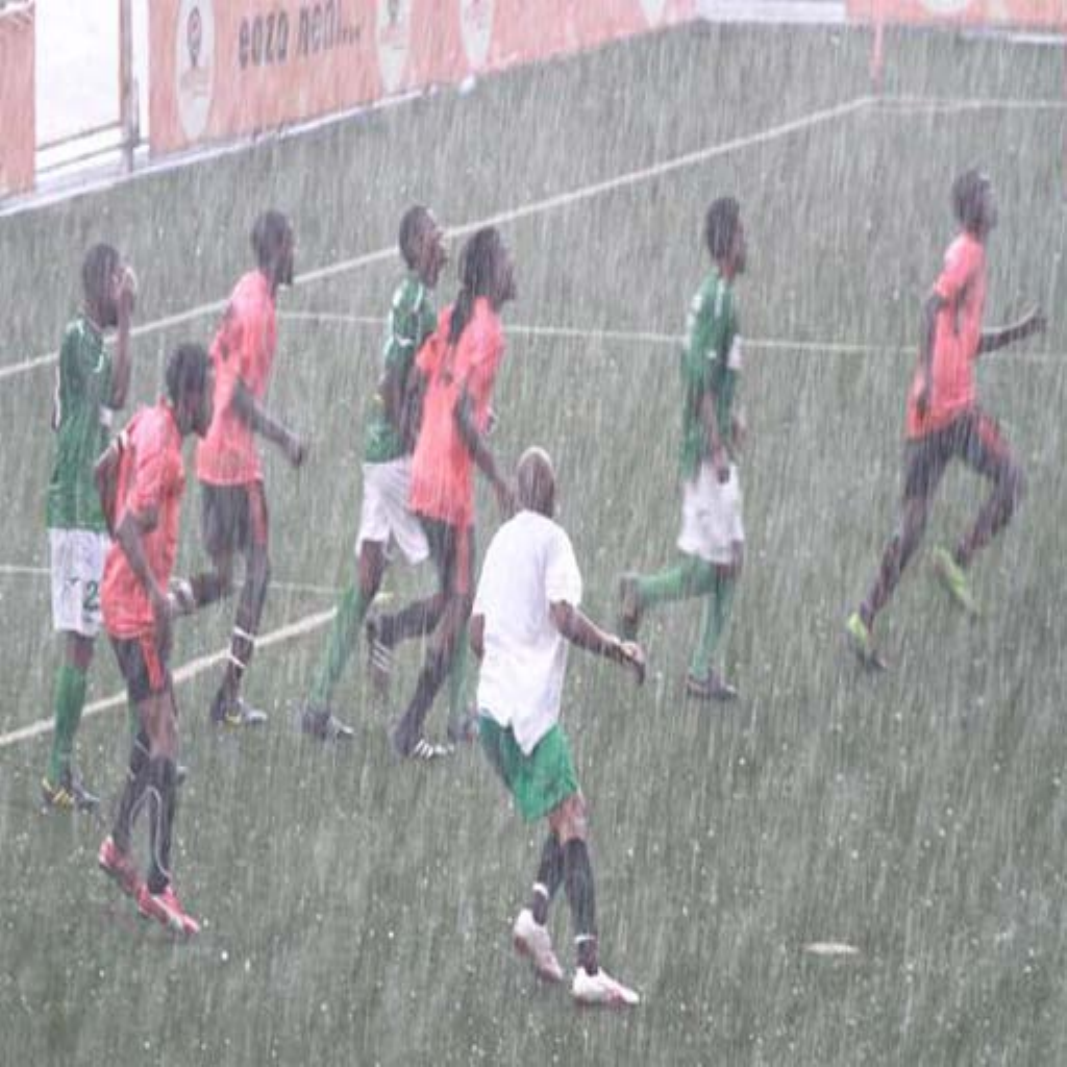}& \hspace{-4mm}
\includegraphics[width = 0.32\linewidth]{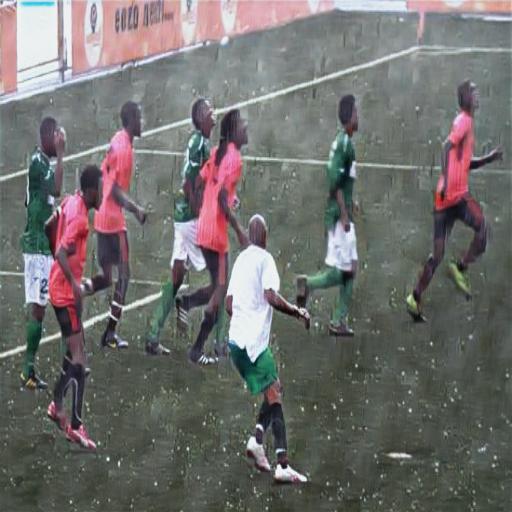}& \hspace{-4mm}
\includegraphics[width = 0.32\linewidth]{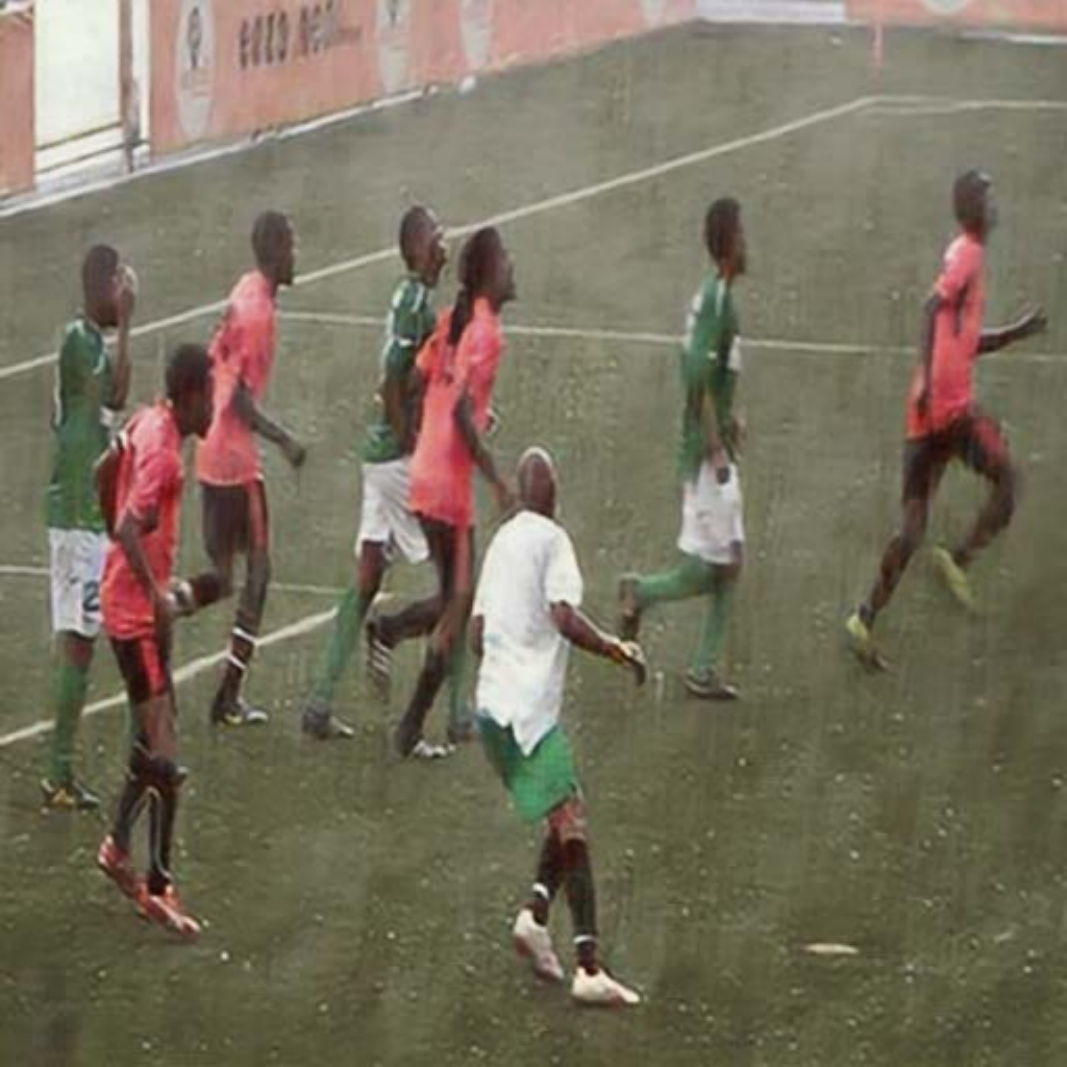}\\
(a) Rainy image &\hspace{-4mm}  (b) GAN for deraining~\cite{derain/gan} &\hspace{-4mm}  (c) Ours \\
\includegraphics[width = 0.32\linewidth]{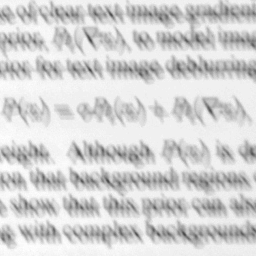} & \hspace{-4mm}
\includegraphics[width = 0.32\linewidth]{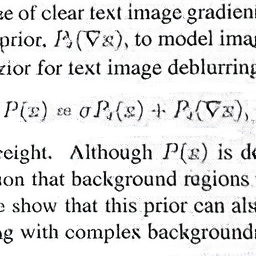} & \hspace{-4mm}
\includegraphics[width = 0.32\linewidth]{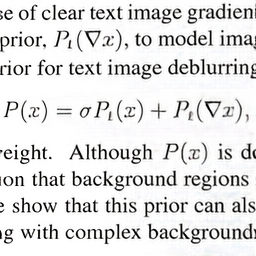}\\
(d) Blurred image &\hspace{-4mm}  (e) GAN for deblurring~\cite{pixel/to/pixel} &\hspace{-4mm}  (f) Ours\\
\end{tabular}
\end{center}
\vspace{-3mm}
\caption{Two image restoration problems.
Our method is motivated by a key observation that the restored results should be consistent with the observed inputs under the degradation process (i.e., physics model).
Enforcing this fundamental constraint in the generative adversarial network (GAN) can generate better results than those
without using the constraint (e.g., (b)).
We show that the proposed method can be applied to several image restoration and low-level vision problems and performs favorably against state-of-the-art algorithms.
}
\label{fig:teaser}
\vspace{-1mm}
\end{figure*}

\vspace{-2mm}
\section{Introduction}
\vspace{-1mm}
\label{sec: introduction}
Numerous image restoration and low-level vision problems (e.g., image deblurring, image super-resolution, image dehazing, and image deraining) aim to estimate a clear image $x$ from a given input $y$.
The fundamental assumption is that the estimated $x$ should be consistent with the input $y$ under the image formation model\footnote{The image formation process is modeled by \eqref{eq: mapping-function-low-level}. In this paper, we refer to \eqref{eq: mapping-function-low-level} as the image degradation model or physics model.}:
\begin{equation}
y = \mathcal{H}(x),
\label{eq: mapping-function-low-level}
\end{equation}
where the operator $\mathcal{H}$ maps the unknown result $x$ to the observed image $y$.
For example, $\mathcal{H}$ corresponds to the blur operation if \eqref{eq: mapping-function-low-level} describes an image deblurring problem.
As estimating $x$ from $y$ is ill-posed, it is necessary to introduce additional constraints to regularize $x$.
One widely-used approach is based on the maximum a posterior (MAP) framework, where $x$ can be solved by
\begin{equation}
x^* = \arg\max_x p(x|y) = \arg\max_x p(y|x)p(x).
\label{eq: map-low-level}
\end{equation}
In the above equation, $p(y|x)$ and $p(x)$ are probability density functions, which are usually referred to as the likelihood term and image prior.
In recent years, several deep models have been developed to deal with image restoration and related low-level vision tasks, e.g., image super-resolution~\cite{SRCNN/pami16,fast/SRCNN/eccv16,SRCNN/eccv14,VDSR/cvpr16,Recurrent/SR/cvpr16,videoSR/iccv15}, image filtering~\cite{deepfilter/icml15,rnnfilter/eccv16},
noise removal~\cite{dong/artifacts/iccv15,denoising/deep/learning/nips08,denoising/deep/learning/nips12}, image deraining~\cite{Eigen/derain}, and dehazing~\cite{dehaze/eccv16,DehazeNet/tip16,AOD/dehazing}, to name a few.
Mathematically, these methods directly learn the mapping functions between $x$ and $y$ with
\begin{equation}
\vspace{-1mm}
x^* = \mathcal{G}(y),
\label{eq: mapping-function}
\vspace{-1mm}
\end{equation}
where $\mathcal{G}$ is the mapping function.
The function $\mathcal{G}$ can be regarded as an inverse operator of $\mathcal{H}$ in~\eqref{eq: mapping-function-low-level}.
Theoretically, $\mathcal{G}(y)$ should be close to the ground truth if the mapping function can be estimated well.
However, due to the complexity of the problem, e.g., the solution space of the corresponding problem is large, a simple model with a random initialization is not sufficient
to estimate the mapping function well.
Thus, only using a feed-forward network to learn the inverse operator $\mathcal{G}$ does not generate good results as we will demonstrate below.

Recently, generative adversarial networks (GANs)~\cite{GAN}
have been applied to image restoration problems, e.g., image super-resolution~\cite{SRGAN}, image deraining~\cite{derain/gan}, and image deblurring~\cite{DeblurGAN}.
The GAN framework contains a generative model and a discriminative model, where the discriminative model is used to regularize the generative model,
such that the distribution of outputs is close to that of realistic images.
However, the adversarial loss does not ensure that the contents of outputs are consistent with those of the inputs.
Although several algorithms~\cite{derain/gan,SRGAN} use a pixel-wise loss function based on the ground truths,
and a perceptual loss function~\cite{Perceptual/Loss} based on
pre-trained VGG features as the constraints in the GAN formulation, these algorithms still do not perform well in image restoration as shown in Figure~\ref{fig:teaser}(b).

We note that the aforementioned methods only aim to estimate the mapping function~\eqref{eq: mapping-function} but do not guarantee whether the
solutions satisfy the physics model~\eqref{eq: mapping-function-low-level} or not.
Without the physics model constraint, the methods based on the feed-forward networks do not restore correct results, e.g., main structures and details of the generated images are incorrect as shown in Figure~\ref{fig:teaser}(b) and (e).
Thus, it is important to develop an algorithm that can model both the  mapping function (i.e.,~\eqref{eq: mapping-function}) and physics model (i.e.,~\eqref{eq: mapping-function-low-level}) in a unified framework to address image restoration and related problems.

In this paper, we propose a GAN model constrained by a physics model for image restoration and low-level vision tasks.
The physics model ensures that the estimated result (i.e., $\mathcal{G}(y)$) should be consistent with the observed image $y$.
The proposed physics constrained GAN model is jointly trained in an end-to-end fashion.
We show that the proposed algorithm can be effectively applied to a variety of image restoration and low-level vision problems (see Figure~\ref{fig:teaser}).

\vspace{-3mm}
\section{Related Work}
\vspace{-1mm}
In this section, we review the methods closely to this work in proper context.
\vspace{-2mm}
{\flushleft \bf{Generative adversarial networks. }}
Goodfellow et al.~\cite{GAN} propose the GAN framework to generate realistic images from random noise.
Motivated by this framework, numerous methods~\cite{pixel/to/pixel,CycleGAN2017,DiscoGAN,dualGAN,LSGAN} have been proposed for vision tasks.
Recently, the GAN framework has also been applied to low-level vision problems~\cite{derain/gan,SRGAN,DeblurGAN,gan/map,Xu_2017_ICCV}.
In this work, we propose an efficient algorithm constrained by the physics model to improve the GAN framework for image deblurring, image dehazing, and related tasks.
\vspace{-6mm}
{\flushleft \bf{Image deblurring. }}
Numerous methods based on statistical priors have been developed to address this ill-posed problem.
In contrast to conventional methods, Schuler et al.~\cite{mlp/non/blind} develop a multi-layer perceptron approach to remove noise and artifacts in the deblurring process.
Xu et al.~\cite{xu/nips/cnn/deconvolution} develop a convolutional neural network based on the singular value decomposition to deal with outliers.
As these methods are designed for non-blind image deblurring, it is not clear how these approaches can be extended to blind image deblurring.
For blind image deblurring, some approaches~\cite{cnn/jiansun/cvpr15,learning/to/deblur/pami} first use convolutional neural networks to estimate blur kernels and then deblur images with the conventional image restoration methods.
To directly restore clear images, several end-to-end trainable neural networks~\cite{CNN/text/deblur,CNN/dynamic/cvpr17} have been developed.
Although these methods alleviate the complex blur kernel estimation step, the generated results usually do not satisfy the image formation model and the structures of the recovered images are not preserved well as shown in Figure~\ref{fig:teaser}(e).
\vspace{-2mm}
{\flushleft \bf{Image dehazing.}}
The success of the conventional dehazing methods is due to effective design of handcrafted features for estimating transmission maps, e.g., dark channel~\cite{he/dark/channel/dehazing/cvpr09}.
Recent deep learning-based methods~\cite{dehaze/eccv16,DehazeNet/tip16} first use neural networks to estimate transmission maps and then restore clear images based on the traditional schemes~\cite{he/dark/channel/dehazing/cvpr09}.
Different from these methods, we propose an end-to-end trainable network to solve the image dehazing problem, which can directly restore clear images from hazy inputs.

\vspace{-2mm}
{\flushleft \bf{Image deraining.}} For the image deraining, conventional algorithms are usually developed based on the statistical properties of rainy streaks~\cite{sparse/coding/deraing/tip12,jihui/iccv15,derain/iccv/ChenH13,liyu/derain/cvpr16}.
Recently, Eigen et al.~\cite{Eigen/derain} develop a neural network to remove rain/dirt in images.
Motivated by the success of the ResNet~\cite{ResNet/CVPR16}, Fu et al.~\cite{derain/cvpr17/fu} develop a deep
network for image deraining.
On the other hand, Zhang et al. develop a GAN-based method~\cite{derain/gan} with a perceptual loss function~\cite{Perceptual/Loss} and
Yang et al.~\cite{Yang_2017_CVPR} develop a multi-task network for rain detection and removal.
As the physical formation model is not considered, these deep models do not effectively solve the deraining problem (see Figure~\ref{fig:teaser}(b)).

\vspace{-2mm}
{\flushleft \bf{Image super-resolution and low-level vision tasks.}}
Significant progress has been made in super-resolution due to the use of deep network models~\cite{SRCNN/eccv14,VDSR/cvpr16,SRGAN,EnhanceNet}.
In~\cite{SRCNN/eccv14}, Dong et al. develop an end-to-end trainable network for super-resolution (SRCNN).
As the SRCNN algorithm is less effective in restoring image details, Kim et al.~\cite{VDSR/cvpr16} propose a deeper network with residual learning.
To generate more realistic images, Ledig et al.~\cite{SRGAN} develop a GAN for image super-resolution.
In addition, the deep learning methods have been applied to other low-level vision problems, such as image filtering~\cite{deepfilter/icml15,rnnfilter/eccv16}
and image denoising~\cite{dong/artifacts/iccv15,denoising/deep/learning/nips08,denoising/deep/learning/nips12}.
Different from these approaches, we propose a GAN-based method constrained by the physics model for image restoration and low-level vision problems.

\vspace{-3mm}
\section{Image Restoration with GAN}
\vspace{-1mm}
The GAN algorithm learns a generative model via an adversarial learning process.
It simultaneously trains a generative network and a discriminative network by optimizing
\vspace{-1mm}
\begin{equation}\label{eq:origin-gan}
\!\!\!\!\min_\mathcal{G}\max_\mathcal{D} \mathbb{E}_{x \!\sim\! p_{\textrm{data}}(x)} [ \log \mathcal{D}(x)]\!+\! \mathbb{E}_{z \!\sim\! p_z(z)} [ \log (1 \!-\! \mathcal{D}(\mathcal{G}(z)))],\!\!\!\!
\vspace{-1mm}
\end{equation}
where $z$ denotes random noise, $x$ is a real image, and $\mathcal{D}$ represents a discriminative network.
For simplicity, we also use $\mathcal{G}$ to denote a generative network.

In the training process, the generator draws samples (i.e., $\mathcal{G}(z)$) that can fool the discriminator,
while the discriminator learns to distinguish the real data and those from the generator.
The discriminator is a binary classifier.
If we use the observed image $y$ as the input of the generator, the adversarial loss is
\vspace{-1mm}
\begin{equation}\label{eq:origin-gan-adversarial}
%
\max_\mathcal{D} \mathbb{E}_{x \sim p_{\textrm{data}}(x)} [ \log \mathcal{D}(x)]+ \mathbb{E}_{y \sim_{\textrm{data}}(y)} [ \log (1 - \mathcal{D}(\mathcal{G}(y)))].
\vspace{-1mm}
\end{equation}
Thus, the value of~\eqref{eq:origin-gan-adversarial} is close to zero if the distribution of the generated image $\mathcal{G}(y)$ is significantly different from that of the clear one and be larger otherwise.

Taking the negative log operation on~\eqref{eq: map-low-level}, the image restoration problem~\eqref{eq: map-low-level} can be solved by
\vspace{-1mm}
\begin{equation}
x^*  = \arg\min_{x} \rho(x, y) + \varphi(x),
\label{eq: map-low-level-energy}
\vspace{-1mm}
\end{equation}
where $\rho(x, y)$ is the data term to ensure that the recovered image $x$ is consistent with the input image $y$ under the corresponding image degradation model~\eqref{eq: mapping-function-low-level},
and $\varphi(x)$ is the regularization of $x$ modeling the properties of $x$ (e.g., sparse gradient distribution~\cite{Levin/CVPR2009}).

For vision tasks, e.g., image deblurring, $\varphi(x)$ acts as a discriminator and the value is much smaller if $x$ is clear and larger otherwise~\cite{jinshan/dark/channel}.
In other words, optimizing the objective function~\eqref{eq: map-low-level-energy} will make the value of $\varphi(x)$ smaller.
Thus, the estimated intermediate image will be much clearer.
Motivated by this, $\varphi(x)$ can be achieved by a binary classifier, which has been used in image deblurring~\cite{cnnprior/lilerenhan}.
Therefore, the adversarial loss can be used as a prior to regularize the solution space of image restoration as demonstrated by~\cite{cnnprior/lilerenhan}.

We note that GAN models with the observed data $y$ as the input have shown promising results in image super-resolution~\cite{SRGAN}, image deraining~\cite{derain/gan}, and image deblurring~\cite{DeblurGAN,Xu_2017_ICCV}.
However, unconstrained GAN models do not guarantee whether the solutions satisfy the image degradation model~\eqref{eq: mapping-function-low-level} or not
(i.e., a GAN model does not consider the effect of the data term in~\eqref{eq: map-low-level-energy}) and thus does not generate clear images as illustrated in Section~\ref{sec: introduction}.
In the following, we propose a physics-based GAN model (i.e., based on~\eqref{eq: mapping-function-low-level}) for image restoration and low-level vision tasks.

\begin{figure*}[t]\footnotesize
\centering
\begin{tabular}{c}
\includegraphics[width = 0.96\linewidth]{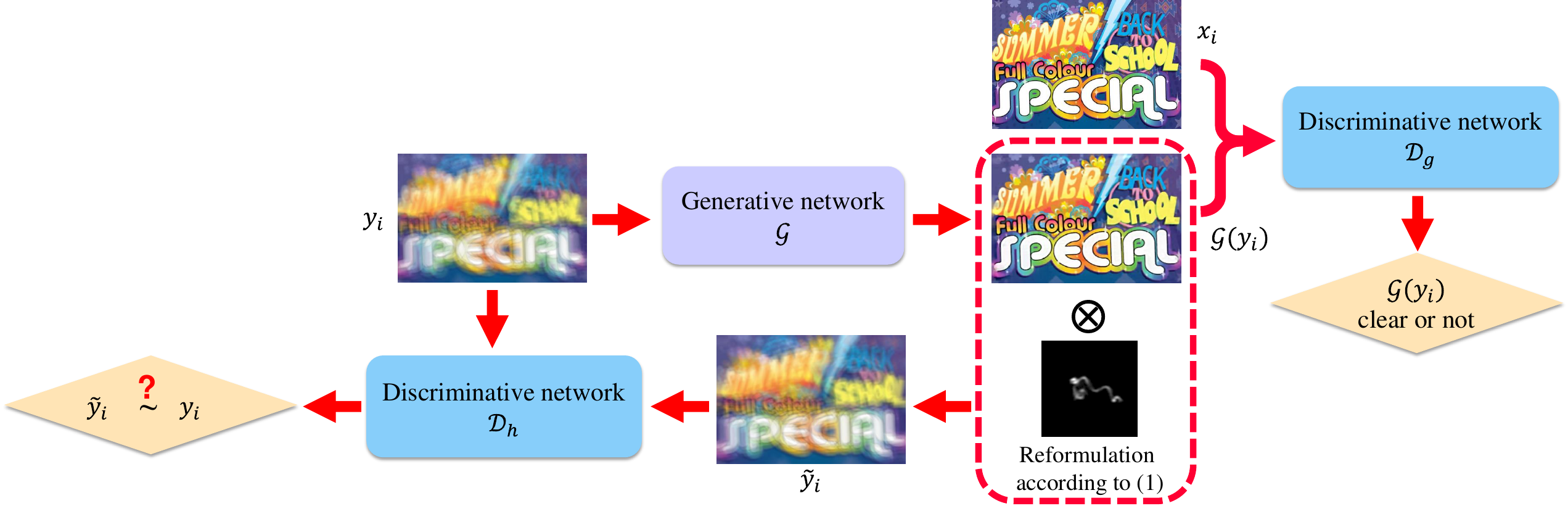}\\
\end{tabular}
\vspace{-2mm}
\caption{Proposed algorithm.
The discriminative network $\mathcal{D}_g$ is used to classify whether the distributions of the outputs from the generator $\mathcal{G}$ are close to those of the ground truth images or not. The discriminative network $\mathcal{D}_h$ is used to classify whether the regenerated result $\widetilde{y}_i$ is consistent with the observed image $y_i$ or not.
All the networks are jointly trained in an end-to-end manner.
}
\label{fig: proposed-framework}
\vspace{-2mm}
\end{figure*}

\vspace{-3mm}
\section{Proposed Algorithm}
\label{sec: proposed-algorithm}
\vspace{-1mm}
To ensure the output of a GAN (i.e., $\mathcal{G}(y)$) is consistent with the input $y$ based on the model~\eqref{eq: mapping-function-low-level}, we introduce an additional discriminative network.
Figure~\ref{fig: proposed-framework} shows the main modules of the proposed algorithm:
two discriminative networks, one generative network, and one image degradation model (i.e.,~\eqref{eq: mapping-function-low-level}).
We use the image deblurring problem as an example.
Let $\{x_{i}\}_{i=1}^{N}$ and $\{y_{i}\}_{i=1}^{N}$ denote the clear and corresponding blurred images.
The generative network learns the mapping function $\mathcal{G}$ and generates the intermediate deblurred image $\mathcal{G}(y_{i})$ from the input $y_{i}$.
Then, we apply the image degradation model~\eqref{eq: mapping-function-low-level} to $\mathcal{G}(y_{i})$
\begin{equation}
\widetilde{y}_{i}  = k_i \otimes \mathcal{G}(y_{i}),
\label{eq: blur-model}
\end{equation}
where $k_i$ denotes a blur kernel used only in the training process\footnote{Note that the blur kernel $k_i$ in~\eqref{eq: blur-model} is known, which is also used to generate the blurred image $y_i$ from the clear image $x_i$ when synthesizing the training data. Thus, the physics model used in the proposed algorithm is not stochastic.}, and $\otimes$ denotes the convolution operator.
The discriminative network $\mathcal{D}_h$ takes the blurred image $y_{i}$ and the regenerated image $\widetilde{y}_{i}$
as the inputs to classify whether the generated results satisfy the blur model or not.
The other discriminative network $\mathcal{D}_g$ takes the ground truth $x_{i}$ and the intermediate deblurred image
$\mathcal{G}(y_i)$ as the inputs to classify whether $\mathcal{G}(x_i)$ is clear or not.
For image dehazing, the physics model is $\widetilde{y}_i = \mathcal{G}(x_i)t_i + A_i(1-t_i)$, where $t_i$ is the transmission map and $A_i$ is the atmospheric light.
In image super-resolution, the physics model is $\widetilde{y}_i = \mathcal{B}(\mathcal{G}(x_i))$, where $\mathcal{B}$ denotes the down-sampling and filtering operator.
In this paper, we use the bicubic interpolation operation.
For other applications, we can use the corresponding physics models for different vision tasks.

We note that although the proposed network is trained in an end-to-end manner, it is constrained by a physics model and thus is not fully blind in the training stage.
With the learned generator $\mathcal{G}$, the test stage is blind.
We can directly obtain the final results by applying it to the input images.

\vspace{-4mm}
\subsection{Network Architecture}
\label{ssec: Network Architecture}
\vspace{-1mm}
Once the physics model constraint is determined for each vision task,
we can use existing network architectures for the generative and discriminative modules.

\vspace{-2mm}
{\flushleft \bf{Generative network.}} The generative network $\mathcal{G}$ is used to render the final results.
Numerous generative networks have been developed for image restoration and low-level vision problems
(e.g., super-resolution~\cite{Perceptual/Loss,SRGAN} and image editing~\cite{CycleGAN2017}).
In this work, we use network model similar to that of the CycleGAN~\cite{CycleGAN2017} as our generative module.
Table~\ref{tab: network-parameters} shows the network parameters.

\vspace{-2mm}
{\flushleft \bf{Discriminative network.}}
The PatchGANs~\cite{SRGAN,CycleGAN2017,pixel/to/pixel} have fewer parameters than a full image discriminator and achieve state-of-the-art results in many vision problems.
In this work, the discriminative models $\mathcal{D}_h$ and $\mathcal{D}_g$ are similar to those of the PatchGANs~\cite{pixel/to/pixel,SRGAN,CycleGAN2017}.
We use the same network parameters for both $\mathcal{D}_h$ and $\mathcal{D}_g$.
The discriminator model parameters are shown in Table~\ref{tab: network-parameters}.

\begin{table*}[!t]
\caption{\label{tab: network-parameters} Network configurations.
``CIR" denotes the convolutional layer with the instance normalization (IN)~\cite{instance/normalization} and ReLU;
``ResBlock" represents the residual block~\cite{ResNet/CVPR16} which contains two convolutional layers with the IN and ReLU;
``CTIR" denotes the transposed convolutional layer with the IN and ReLU;
``CLR" denotes the convolutional layer with LeakyReLU;
``CILR" denotes the convolutional layer with IN and LeakyReLU; ``CT" denotes the convolutional layer with the Tanh function; ``C" denotes the convolutional layer.
}
\vspace{-3mm}
\centering
\begin{tabular}{lccc|cc|ccc}
\multicolumn{9}{c}{Parameters of the generative network}           \\
\toprule
Layers                 & CIR$_1$ & CIR$_2$ & CIR$_3$ & \multicolumn{2}{c|}{ResBlock$_1$-ResBlock$_9$} & CTIR$_1$ & CTIR$_2$ & CT \\
\hline
Filter size            &  7  &  3  & 3           & 3              & 3                       &  3   & 3    &  7   \\
Filter numbers         & 64  & 128 & 256         & ~~~~~256~~~~~  &~~~~~256~~~~~            & 128  & 64   &  3   \\
Stride                 & 1   & 2    & 2          & 1              & 1                       & 2  & 2  &  1    \\
\bottomrule
\vspace{-1mm}
\end{tabular}
\begin{tabular}{lccccc}
\multicolumn{6}{c}{Parameters of the discriminative network}           \\
\toprule
Layers                 & CLR & CILR$_1$ & CILR$_2$ & CILR$_3$ & C \\
\hline
Filter size            &  4  &  4  & 4           & 4               &  4   \\
Filter numbers         & 64  & 128 & 256         & 512             & 1 \\
Stride                 & 2   & 2    & 2          & 1               & 1 \\
\bottomrule
\end{tabular}
\vspace{-2mm}
\end{table*}

\vspace{-3mm}
\subsection{Loss Function}
\vspace{-1mm}
A straightforward GAN model for image restoration is to use~\eqref{eq:origin-gan}.
However, the contents of the generated images based on this training loss may be different from the ground truth images as demonstrated by~\cite{SRGAN}.
To ensure that the contents of the generated results from the generative network are close to those of the ground truth images and also consistent with those of the inputs under the physics model~\eqref{eq: mapping-function-low-level},
we use the $L_1$ norm regularized pixel-wise loss functions
\begin{equation}
\mathcal{L}_p = \sum_i\|\widetilde{y}_i - y_i\|_1,
\label{eq: pixel-wise-loss-co}
\end{equation}
and
\begin{equation}
\mathcal{L}_g = \sum_i\|\mathcal{G}(y_i) - x_i\|_1,
\label{eq: pixel-wise-loss}
\end{equation}
in the training stage.
To make the generative network learning process more stable, we use the loss function
\vspace{-1mm}
\begin{equation}
\mathcal{\widetilde{L}}_g = \sum_i\|\mathcal{G}(\widetilde{y}_i) - x_{i}\|_1,
\label{eq: pixel-wise-loss-co-2}
\vspace{-1mm}
\end{equation}
to regularize the generator $\mathcal{G}$.

In this work, we propose the objective function
\vspace{-1mm}
\begin{equation}
\begin{split}
\label{eq: gan-loss-energy}
\mathcal{L}_a = \sum_i[\log(\mathcal{D}_g(x_{i}))] + [\log(1-\mathcal{D}_g(\mathcal{G}(y_i)))] \\+[\log(\mathcal{D}_h(y_{i}))] + [\log(1-\mathcal{D}_h(\widetilde{y}_i))],
\end{split}
\end{equation}
to ensure that the output of the proposed GAN model is consistent with the observed input under the image formation model~\eqref{eq: mapping-function-low-level}.

Based on the above considerations, the $\mathcal{G}$, $\mathcal{D}_g$, and $\mathcal{D}_h$ models are trained by solving
\begin{equation}
(\mathcal{G}^*, \mathcal{D}_g^*, \mathcal{D}_h^*) = \min_\mathcal{G}\max_\mathcal{D} \frac{1}{N}\left\{\mathcal{L}_a + \lambda \mathcal{L}_g + \gamma\mathcal{L}_p + \mu\mathcal{\widetilde{L}}_g\right\},
\label{eq: loss-for-all}
\end{equation}
where $\lambda$, $\gamma$, and $\mu$ are weight parameters.
To make the training process stable, we use the least square GAN loss~\cite{LSGAN} in $\mathcal{L}_a$ in this work.

\vspace{-3mm}
\section{Experimental Results}
\vspace{-1mm}
In this section, we evaluate the proposed algorithm on several image restoration tasks including image deblurring, dehazing, super-resolution, and deraining.
We show the main experimental results in the main paper and present more  analysis as well as applications in the supplementary material.
The trained models, datasets, source code, and supplementary material are available at \url{https://jspan.github.io/projects/physicsgan/}.

\vspace{-2mm}
\subsection{Datasets}
\vspace{-1mm}
\label{ssec: datasets}
For image deblurring, we use the training dataset by Hradi{\v{s}}~\etal~\cite{CNN/text/deblur}, which consists of images with both defocus blur generated by anti-aliased disc and motion blur synthesized based on random walk.
We randomly crop one million $256\times 256$ blurred image patches from the dataset for training and use the test dataset~\cite{CNN/text/deblur} which includes 100 clear images
for evaluation.
In addition, with the same blur kernels~\cite{CNN/text/deblur}, we randomly choose 50,000 clear face images from the CelebA dataset~\cite{CelebA} and 50,000 natural images
from the COCO dataset~\cite{coco} to generate the training data for the face image deblurring and natural image deblurring.
We add random noise to each blurred image, where the noise level ranges from 0 to 10\%.

For image dehazing, we use the NYU depth dataset~\cite{nyu/depthdata/eccv12} and select 1,449 clear images and corresponding depth maps to generate hazy images according to the hazy model~\cite{he/dark/channel/dehazing/cvpr09}.
As the images in the NYU depth dataset~\cite{nyu/depthdata/eccv12} are indoor images, we also randomly choose 964 outdoor images from the Make3D dataset~\cite{make3d} to synthesize outdoor hazy images as the dataset.
Thus, there are 2,413 synthetic images.
All the images are resized to the size of $512\times 512$ pixels.
To evaluate image dehazing methods, we randomly select 241 images as the test dataset and the remaining images are used for training.
In addition to these synthetic images, we also compare with the state-of-the-art methods using
real captured hazy images.

For the image deraining task, we use the dataset by Zhang et al.~\cite{derain/gan} to train and evaluate the proposed algorithm against the state-of-the-art methods.
\vspace{-3mm}
\subsection{Training}
\vspace{-1mm}
We train the models using the Adam optimizer~\cite{Adam} with an initial learning rate of 0.0002,
and the updating strategy of learning rate is the same as~\cite{CycleGAN2017}.
We set the batch size to be 1 and the slope of the LeakyReLU is 0.2.
We use the same weight initialization method as~\cite{CycleGAN2017}.
After obtaining the generator $\mathcal{G}$, as we know the paired training data \{$x_i$,$y_i$\} and  corresponding physics model parameters (e.g., $k_i$ in image deblurring) that are used to synthesize $y_i$ from $x_i$, we apply the same physics model parameters to $\mathcal{G}(y_i)$ and generate $\widetilde{y}_i$.
The discriminator $\mathcal{D}_g$ then takes $x_i$ and $\mathcal{G}(y_i)$ as the input while the discriminator $\mathcal{D}_h$ takes $y_i$ and $\widetilde{y}_i$ as the input.
Similar to~\cite{CycleGAN2017,Adversarial/update}, we update the discriminators using a history of generated images instead of the ones by the latest generative networks according to~\cite{CycleGAN2017}.
The update ratio between the generator and the discriminators is set to be 1.
%
We empirically set $\gamma$, $\mu$, and $\lambda$ to be 10, 10, and 100, respectively.

\begin{table*}[!t]
\caption{\label{tab: psnr-deblurring} Quantitative evaluations with the state-of-the-art methods on the text image deblurring dataset by~\cite{CNN/text/deblur}.
}
\vspace{-3mm}
\centering
\resizebox{0.9\textwidth}{!}{
\begin{tabular}{cccccccccccc}
\toprule
Methods & Input & Xu~\cite{Xu/l0deblur/cvpr2013} & Pan~\cite{jinshan/pami17} & Pan~\cite{jinshan/dark/channel} & CNN~\cite{CNN/text/deblur} & Nah~\cite{CNN/dynamic/cvpr17} & pix2pix~\cite{pixel/to/pixel} & CycleGAN~\cite{CycleGAN2017} & PCycleGAN~\cite{CycleGAN2017} & Ours \\
\midrule
PSNR      & 18.52           & 17.52             & 18.19            & 18.47          & 26.53            & 22.57             & 23.33      & 11.92   &  19.71 & \bf{28.80}\\
SSIM      & 0.6658          & 0.4186            & 0.6270           & 0.6127         & 0.9422           & 0.8924            &0.9170      & 0.2792   & 0.5833 & \bf{0.9744}\\
\bottomrule
\end{tabular}
}
\vspace{-2mm}
\end{table*}

\begin{figure*}[!t]
\centering
\begin{tabular}{cccc}
\includegraphics[width = 0.24\linewidth]{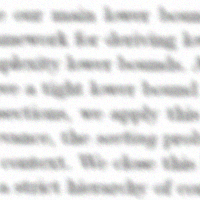}& \hspace{-4mm}
\includegraphics[width = 0.24\linewidth]{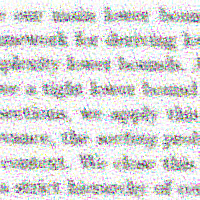}& \hspace{-4mm}
\includegraphics[width = 0.24\linewidth]{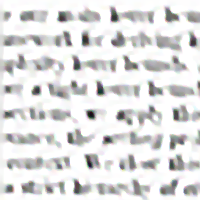}& \hspace{-4mm}
\includegraphics[width = 0.24\linewidth]{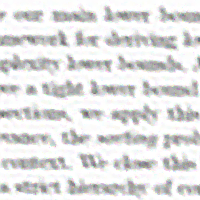}\\
(a) Input & \hspace{-4mm} (b) Xu~\cite{Xu/l0deblur/cvpr2013} & \hspace{-4mm} (c) Pan~\cite{jinshan/pami17} & \hspace{-4mm} (d) Pan~\cite{jinshan/dark/channel} \\
\includegraphics[width = 0.24\linewidth]{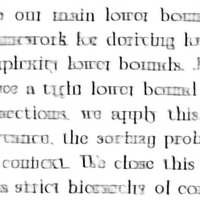} & \hspace{-4mm}
\includegraphics[width = 0.24\linewidth]{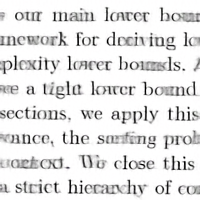}& \hspace{-4mm}
\includegraphics[width = 0.24\linewidth]{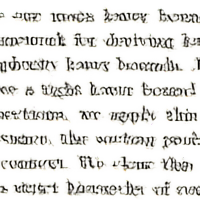}& \hspace{-4mm}
\includegraphics[width = 0.24\linewidth]{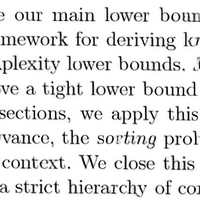}\\
 (e) Nah~\cite{CNN/dynamic/cvpr17} & \hspace{-4mm} (f) CNN~\cite{CNN/text/deblur} & \hspace{-4mm} (g) PCycleGAN~\cite{CycleGAN2017} & \hspace{-4mm} (h) Ours\\
\end{tabular}
\vspace{-1mm}
\caption{One synthetic blurred image from the text image deblurring dataset~\cite{CNN/text/deblur}. The proposed method generates images with much clearer characters.
}
\label{fig: deblur-synthetic}
\vspace{-2mm}
\end{figure*}

\vspace{-3mm}
\subsection{Image Deblurring}
\label{ssec: image-deblurring}
\vspace{-1mm}
We evaluate the proposed deblurring algorithm against the state-of-the-art methods based on
statistical priors~\cite{jinshan/dark/channel,Xu/l0deblur/cvpr2013,jinshan/pami17} and
deep neural networks~\cite{DeblurGAN,CNN/text/deblur,CNN/dynamic/cvpr17,jiawei/cvpr18}.
We note that recent CNN-based deblurring algorithms~\cite{DeblurGAN,CNN/dynamic/cvpr17,jiawei/cvpr18} are designed for natural images.
For deblurring text and face images, we retrain these algorithms using the same training datasets for fair comparisons.
For natural image deblurring, we use the trained models~\cite{DeblurGAN,CNN/dynamic/cvpr17,jiawei/cvpr18} for fair comparisons.
We note that the pix2pix~\cite{pixel/to/pixel} and CycleGAN~\cite{CycleGAN2017} algorithms are designed for image-to-image translation which can be applied to restoration.
We retrain these two algorithms using the same training datasets as the proposed method for fair comparisons.
Since the original CycleGAN algorithm~\cite{CycleGAN2017} is trained with unpaired training data, the CycleGAN model trained with the paired images is referred to as the PCycleGAN method.

\vspace{-2mm}
{\flushleft \bf{Synthetic blurred images.}}
We evaluate the proposed algorithm using the text image dataset described in Section~\ref{ssec: datasets}.
Table~\ref{tab: psnr-deblurring} shows that the proposed algorithm performs favorably against state-of-the-art methods in terms of PSNR and SSIM\footnote{As the implementation of~\cite{textdeblurring/eccv16} is not available, we do not compare this method in this paper. When computing the PSNR and SSIM values, we consider the possible shift that is caused by blur kernels~\cite{jinshan/pami17}.}.
Note that the method by~\cite{CNN/dynamic/cvpr17} uses a multi-scale CNN with adversarial learning.
However, it is less effective than the proposed algorithm as shown in Table~\ref{tab: psnr-deblurring}.
The pix2pix~\cite{pixel/to/pixel} method performs similar as~\cite{CNN/dynamic/cvpr17}
since both schemes are based on GANs with the pixel-wise loss function~\eqref{eq: pixel-wise-loss}.
In contrast, the proposed algorithm exploits a physics model which is effective for image deblurring.

Figure~\ref{fig: deblur-synthetic} shows a blurred text image from the test dataset.
The conventional algorithms~\cite{jinshan/dark/channel,Xu/l0deblur/cvpr2013,jinshan/pami17}
do not generate clear images.
Although the CNN-based method~\cite{CNN/text/deblur} performs better,
the deblurred images contain significant blur residual
as the feed-forward network does not consider the consistency
between the estimated results and blurred inputs.
We note that the CycleGAN method~\cite{CycleGAN2017} uses the cycle consistency constraint for the image-to-image translation task.
However, it does not generate clear images as shown in Figure~\ref{fig: deblur-synthetic}(g) although it is retrained with the paired data.
In contrast, the proposed algorithm is able to generate clearer images with recognizable characters as shown in Figure~\ref{fig: deblur-synthetic}(h), which demonstrates
the importance of the image degradation model constraint.

\begin{table*}[!t]
\caption{\label{tab: psnr-deblurring-face} Quantitative evaluations with the state-of-the-art deblurring methods on face images.
}
\vspace{-3mm}
\centering
\resizebox{0.9\textwidth}{!}{
\begin{tabular}{cccccccccc}
\toprule
Methods  & Xu~\cite{Xu/l0deblur/cvpr2013} & Pan~\cite{jinshan/eccv14} & Pan~\cite{jinshan/dark/channel} & Zhang~\cite{jiawei/cvpr18} & Nah~\cite{CNN/dynamic/cvpr17} & DeblurGAN~\cite{DeblurGAN} & CycleGAN~\cite{CycleGAN2017} &PCycleGAN~\cite{CycleGAN2017} & Ours \\
\midrule
PSNR     & 18.84             & 18.85            &   21.51       &     23.22        &    22.48          & 19.18      & 20.73    &  21.82  & \bf{24.17}\\
SSIM     & 0.4054            & 0.4652           &  0.4263       &     0.6832     &      0.4962       &0.2563      & 0.5978      &  0.6018    & \bf{0.7705}\\
\bottomrule
\end{tabular}
}
\vspace{-2mm}
\end{table*}

\begin{figure*}[!t]
\centering
\begin{tabular}{cccc}
\includegraphics[width = 0.24\linewidth]{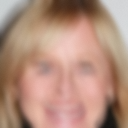}& \hspace{-4mm}
\includegraphics[width = 0.24\linewidth]{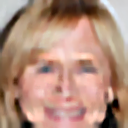}& \hspace{-4mm}
\includegraphics[width = 0.24\linewidth]{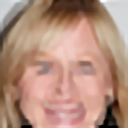}& \hspace{-4mm}
\includegraphics[width = 0.24\linewidth]{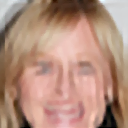}\\
(a) Input & \hspace{-4mm} (b) Xu~\cite{Xu/l0deblur/cvpr2013} & \hspace{-4mm} (c) Pan~\cite{jinshan/eccv14} &\hspace{-4mm}  (d) Pan~\cite{jinshan/dark/channel} \\
\includegraphics[width = 0.24\linewidth]{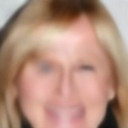} & \hspace{-4mm}
\includegraphics[width = 0.24\linewidth]{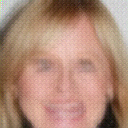}& \hspace{-4mm}
\includegraphics[width = 0.24\linewidth]{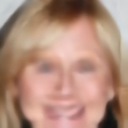}& \hspace{-4mm}
\includegraphics[width = 0.24\linewidth]{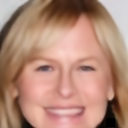}\\
 (e) Zhang~\cite{jiawei/cvpr18} &\hspace{-4mm}  (f) DeblurGAN~\cite{CNN/text/deblur} &\hspace{-4mm} (g) PCycleGAN~\cite{CycleGAN2017} &\hspace{-4mm}  (h) Ours\\
\end{tabular}
\vspace{-1mm}
\caption{Face image deblurring results. The proposed method generates images with fewer artifacts.
}
\label{fig: deblur-face}
\vspace{-1mm}
\end{figure*}

\begin{table*}[!t]
\caption{\label{tab: psnr-deblurring-natural} Quantitative evaluations with the state-of-the-art deblurring methods on natural images.
}
\vspace{-3mm}
\centering
\resizebox{0.9\textwidth}{!}{
\begin{tabular}{cccccccccc}
\toprule
Methods & Xu~\cite{Xu/l0deblur/cvpr2013} & Pan~\cite{jinshan/pami17} & Pan~\cite{jinshan/dark/channel} & Zhang~\cite{CNN/text/deblur} & Nah~\cite{CNN/dynamic/cvpr17} & DeblurGAN~\cite{pixel/to/pixel} &CycleGAN~\cite{CycleGAN2017} & PCycleGAN~\cite{CycleGAN2017} & Ours \\
\midrule
PSNR     & 20.11             & 19.97            & 20.72         &22.48            & 20.89             & 20.10    &  19.98 & 21.86       & \bf{22.63}\\
SSIM     & 0.3802            & 0.4419            & 0.3450       &0.5982            & 0.4878             &0.4596  &   0.5963  & 0.6198       & \bf{0.7151}\\
\bottomrule
\end{tabular}
}
\vspace{-2mm}
\end{table*}


\begin{figure*}[!t]
\centering
\begin{tabular}{cccc}
\includegraphics[width = 0.24\linewidth]{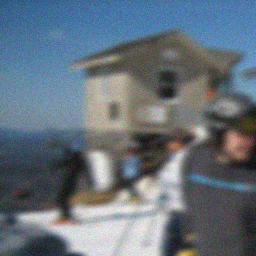}& \hspace{-4mm}
\includegraphics[width = 0.24\linewidth]{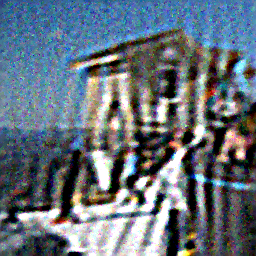}& \hspace{-4mm}
\includegraphics[width = 0.24\linewidth]{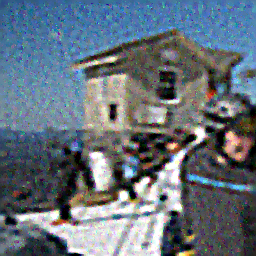}& \hspace{-4mm}
\includegraphics[width = 0.24\linewidth]{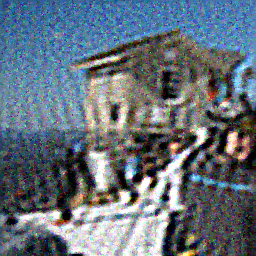}\\
(a) Input & \hspace{-4mm} (b) Xu~\cite{Xu/l0deblur/cvpr2013} & \hspace{-4mm} (c) Pan~\cite{jinshan/pami17} &\hspace{-4mm}  (d) Pan~\cite{jinshan/dark/channel} \\
\includegraphics[width = 0.24\linewidth]{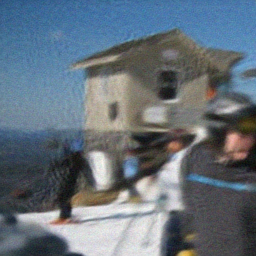} & \hspace{-4mm}
\includegraphics[width = 0.24\linewidth]{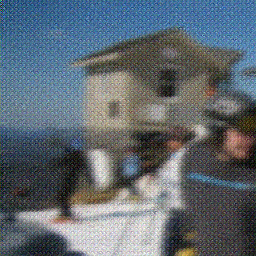}& \hspace{-4mm}
\includegraphics[width = 0.24\linewidth]{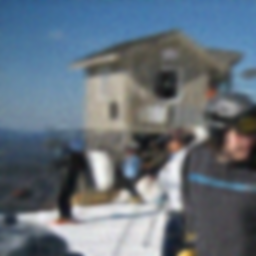}& \hspace{-4mm}
\includegraphics[width = 0.24\linewidth]{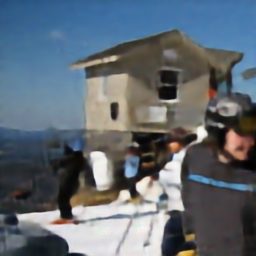}\\
 (e) Zhang~\cite{jiawei/cvpr18} &\hspace{-4mm}  (f) DeblurGAN~\cite{DeblurGAN} &\hspace{-4mm}  (g) PCycleGAN~\cite{CycleGAN2017} &  (h) Ours\\
\end{tabular}
\vspace{-1mm}
\caption{Natural image deblurring results. The proposed method generates images with fewer artifacts.
}
\label{fig: deblur-natural}
\vspace{-1mm}
\end{figure*}

\begin{figure*}[!t]
\centering
\begin{tabular}{cccc}
\includegraphics[width = 0.24\linewidth]{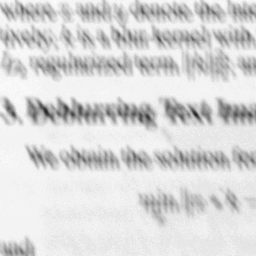}& \hspace{-4mm}
\includegraphics[width = 0.24\linewidth]{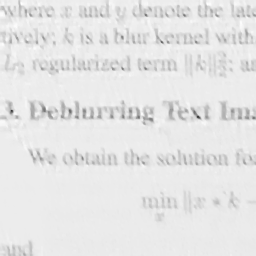}& \hspace{-4mm}
\includegraphics[width = 0.24\linewidth]{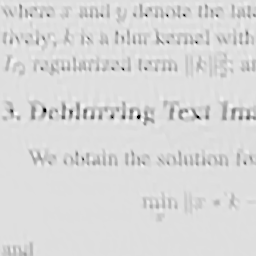}& \hspace{-4mm}
\includegraphics[width = 0.24\linewidth]{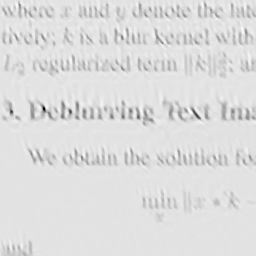}\\
(a) Input & \hspace{-4mm} (b) Xu~\cite{Xu/l0deblur/cvpr2013} & \hspace{-4mm} (c) Pan~\cite{jinshan/pami17} &\hspace{-4mm}  (d) Pan~\cite{jinshan/dark/channel} \\
\includegraphics[width = 0.24\linewidth]{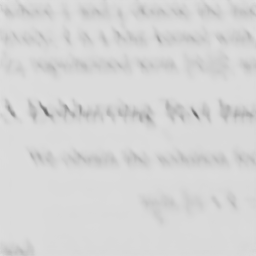} & \hspace{-4mm}
\includegraphics[width = 0.24\linewidth]{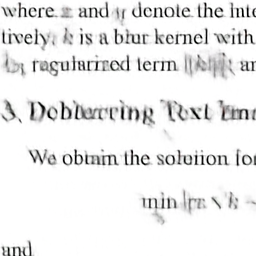}& \hspace{-4mm}
\includegraphics[width = 0.24\linewidth]{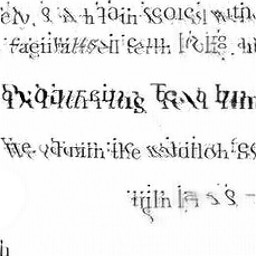}& \hspace{-4mm}
\includegraphics[width = 0.24\linewidth]{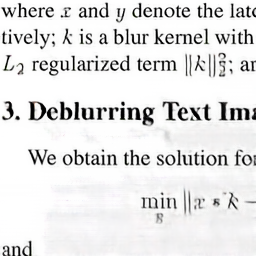}\\
 (e) Nah~\cite{CNN/dynamic/cvpr17} &\hspace{-4mm}  (f) CNN~\cite{CNN/text/deblur} &\hspace{-4mm} (g) PCycleGAN~\cite{CycleGAN2017} &\hspace{-4mm}  (h) Ours\\
\end{tabular}
\caption{Real text image deblurring results. The proposed method generates images with much clearer characters.
}
\label{fig: deblur-real}
\vspace{-1mm}
\end{figure*}

%
To evaluate the proposed method on face image deblurring, we first randomly select 160 clear face images from CelebA dataset~\cite{CelebA} and generate 160 blur kernels according to~\cite{schmidt/cvpr2013/discriminative}, where the clear images and generated blur kernels are not used in the training process.
We then synthesize 160 blurred face images based on clear face images and generated kernels for test.
We compare the proposed algorithm against state-of-the-art methods based on
statistical priors~\cite{jinshan/dark/channel,Xu/l0deblur/cvpr2013,jinshan/pami17}
and deep neural networks~\cite{DeblurGAN,CNN/dynamic/cvpr17,jiawei/cvpr18}.
Table~\ref{tab: psnr-deblurring-face} shows that the proposed algorithm performs favorably against
state-of-the-art methods in terms of PSNR and SSIM and Figure~\ref{fig: deblur-face} shows some deblurred results.
As the input image contains significant blur, several evaluated methods~\cite{CycleGAN2017,CNN/dynamic/cvpr17,jiawei/cvpr18,jinshan/eccv14} do not generate clear images.
We note neither the DeblurGAN nor PCycleGAN method performs well on the face image shown in Figure~\ref{fig: deblur-face}.
In contrast, the proposed algorithm is able to effectively deblur this image which can be attributed to the image degradation model constraint.
We further evaluate the proposed method on natural image deblurring. Similar to the settings used in face image deblurring task,
we randomly select 160 clear images from COCO dataset~\cite{coco} and generate 160 blur kernels according to~\cite{schmidt/cvpr2013/discriminative},
where the clear images and generated blur kernels are not used in the training process.
We synthesize 160 blurred natural images based on clear natural images and generated kernels for test.
The results in Table~\ref{tab: psnr-deblurring-natural} show that the proposed method performs well on natural image deblurring.
In addition, although the proposed network is trained on the uniform blurred images, it also performs well in the dynamic scenes~\cite{CNN/dynamic/cvpr17}
as the blur in dynamic scenes can be approximated by locally uniform blur model~\cite{jinshan/soft/seg/deblur}. More results are presented in the supplementary material.

\vspace{-2mm}
{\flushleft \bf{Real blurred images.}}
Figure~\ref{fig: deblur-real}(a) shows one of the real blurred images in the experiments.
The deblurred image by the proposed method is clearer than those by other algorithms with sharper characters  as shown in Figure~\ref{fig: deblur-real}(h).
More results on real blurred images are presented in the supplementary material.

\begin{table*}[!t]
\caption{\label{tab: psnr-dehazing} Quantitative evaluations with the state-of-the-art methods on the proposed image dehazing dataset.
}
\vspace{-3mm}
\centering
\begin{tabular}{cccccccccc}
\toprule
Methods      & Input       & He~\cite{he/dark/channel/dehazing/cvpr09} & Berman~\cite{non/local/dehazing} & Ren~\cite{dehaze/eccv16} & Cai~\cite{DehazeNet/tip16}& pix2pix~\cite{pixel/to/pixel} & Ours \\
\midrule
PSNR         & 13.10       & 18.01             & 16.67                             & 17.73                    & 20.20          & 24.75       &\bf{24.78}\\
SSIM         & 0.6958      & 0.7784             & 0.7350                           & 0.7719                   & 0.8134         & 0.8225      &\bf{0.8657} \\
\bottomrule
\end{tabular}
\vspace{-2mm}
\end{table*}
\begin{figure*}[!t]
\centering
\begin{tabular}{cccc}
\includegraphics[width = 0.24\linewidth]{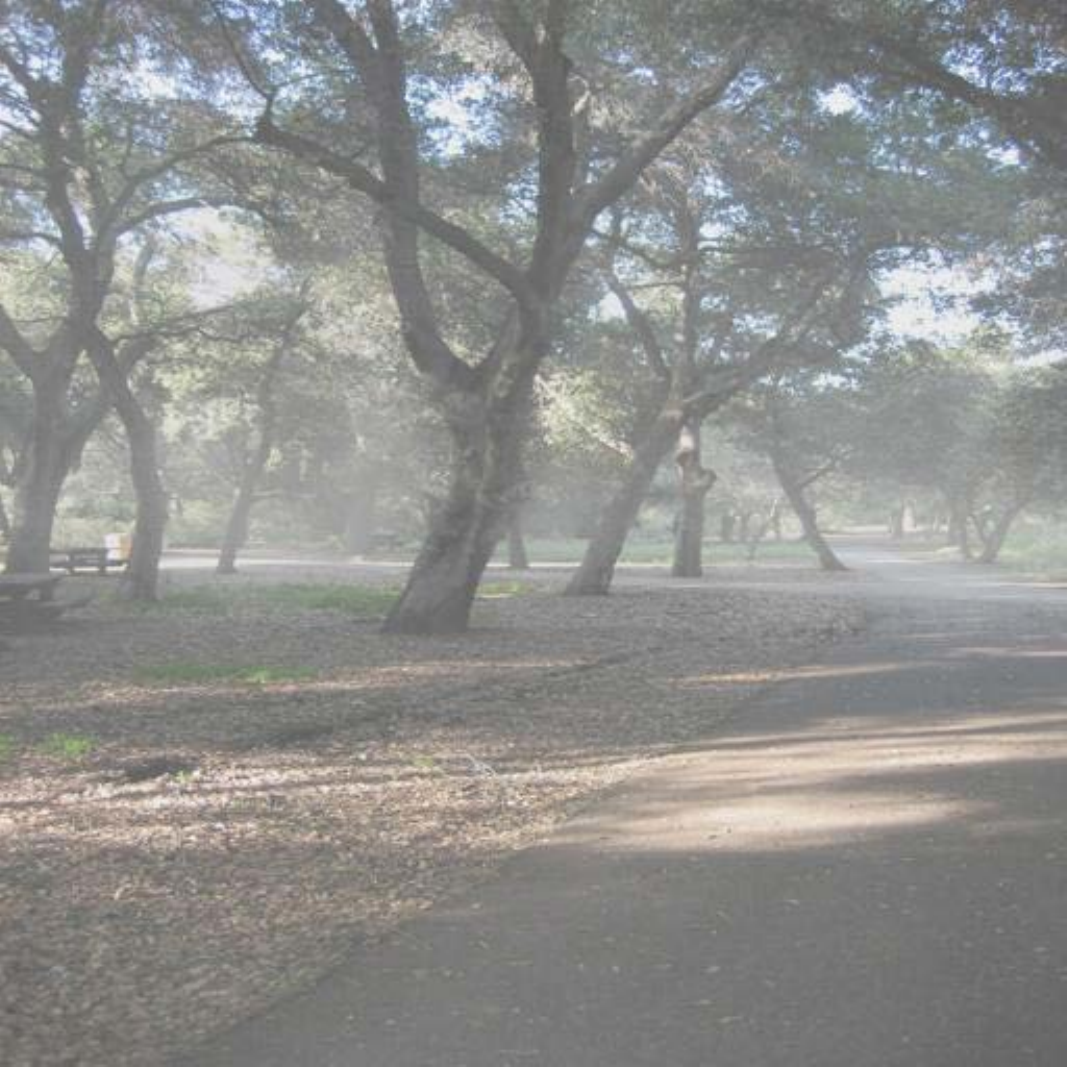}& \hspace{-4mm}
\includegraphics[width = 0.24\linewidth]{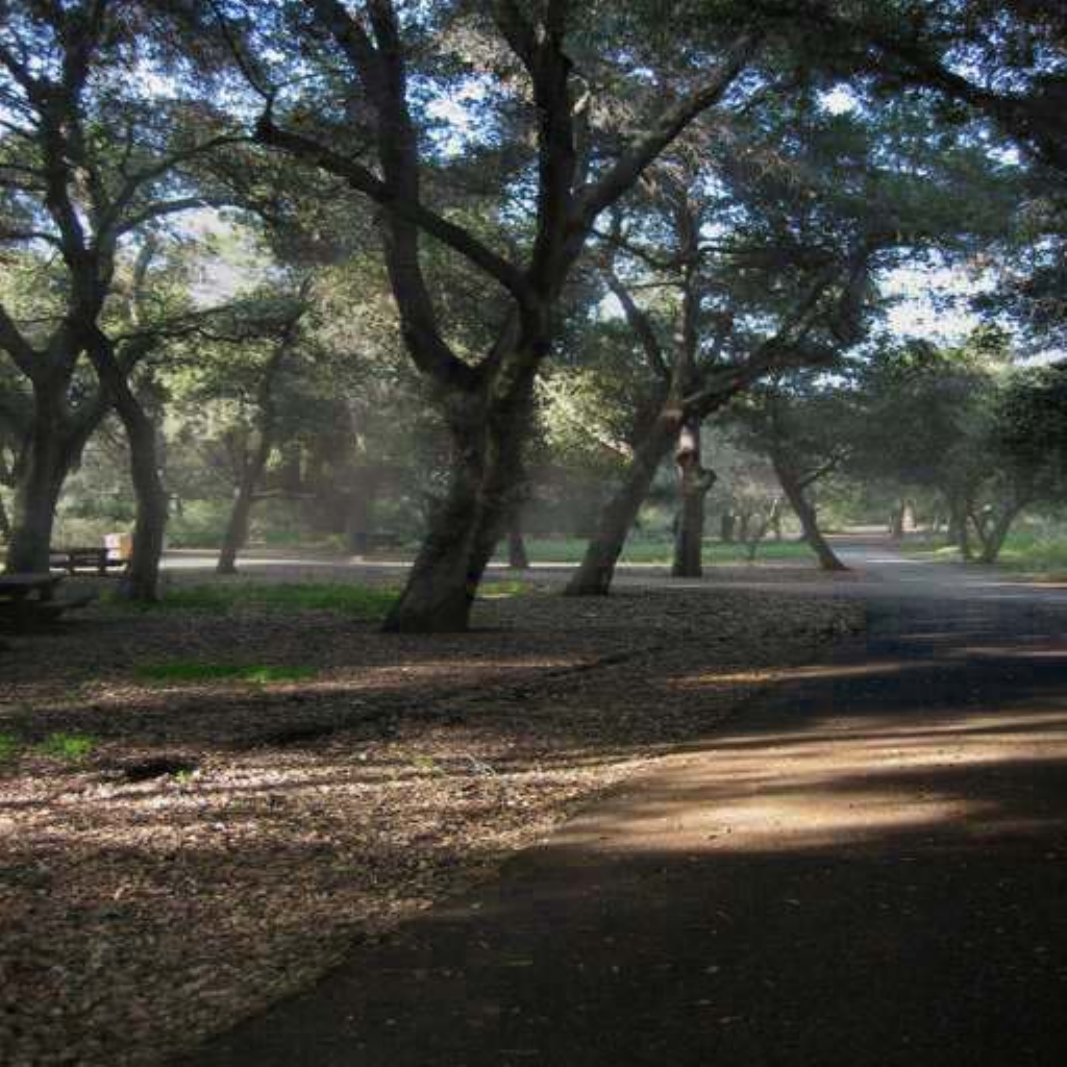}& \hspace{-4mm}
\includegraphics[width = 0.24\linewidth]{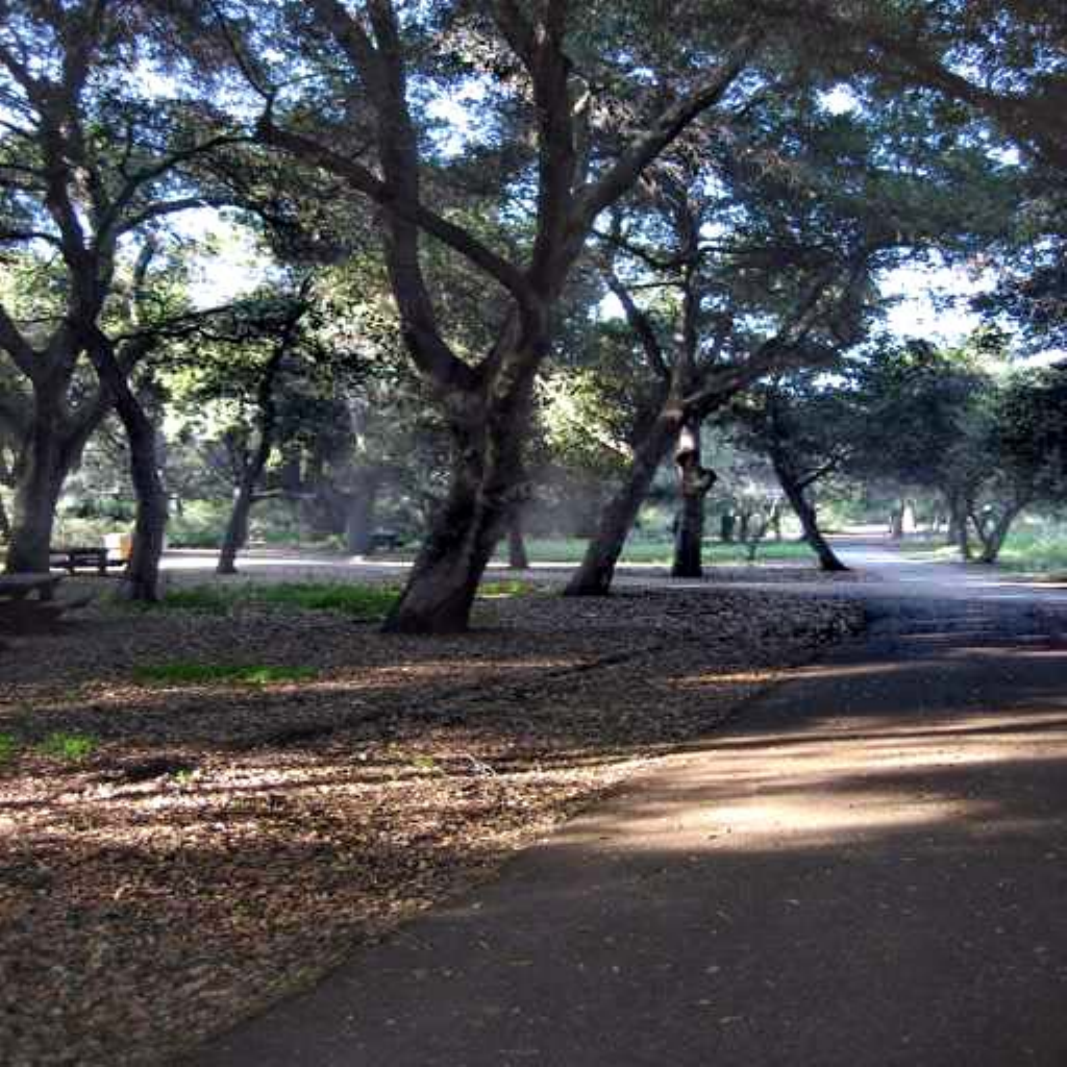}& \hspace{-4mm}
\includegraphics[width = 0.24\linewidth]{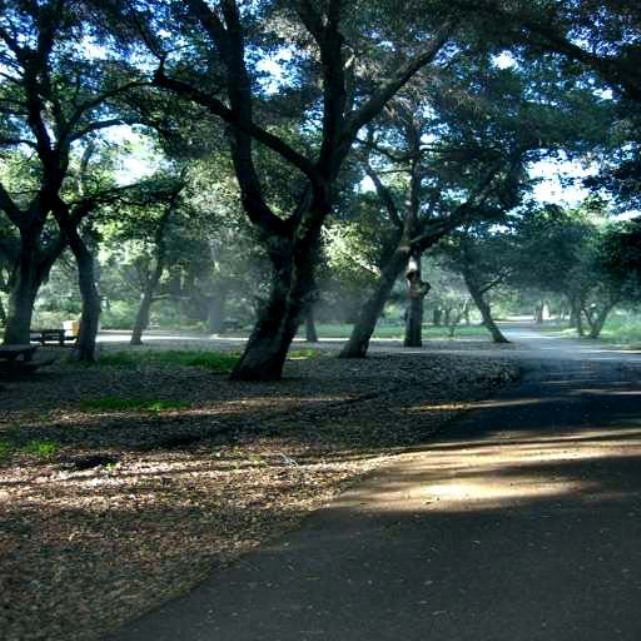}\\
(a) Input &\hspace{-4mm} (b) He ~\cite{he/dark/channel/dehazing/cvpr09} &\hspace{-4mm} (c) Berman~\cite{non/local/dehazing} &\hspace{-4mm} (d) Ren~\cite{dehaze/eccv16}\\
\includegraphics[width = 0.24\linewidth]{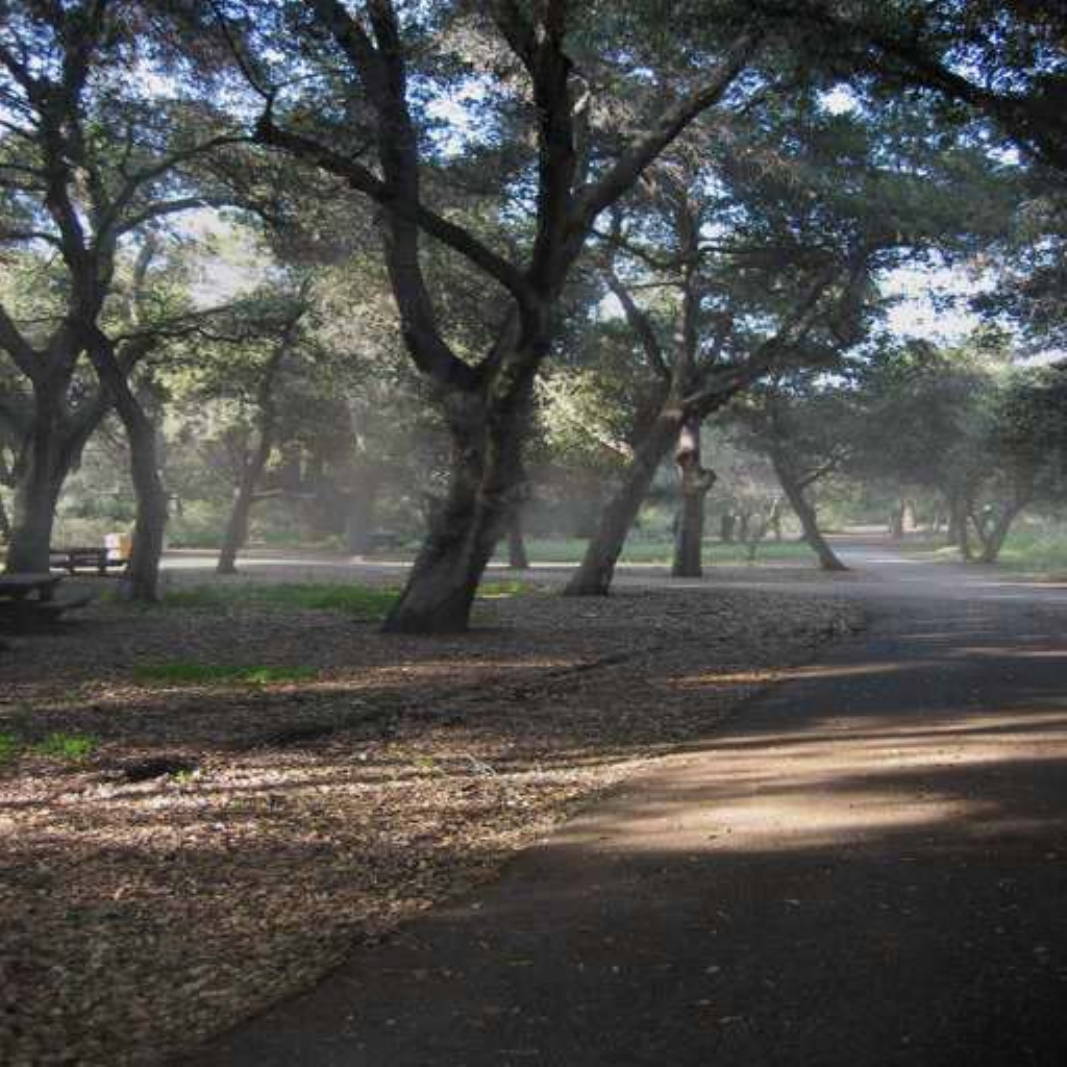}& \hspace{-4mm}
\includegraphics[width = 0.24\linewidth]{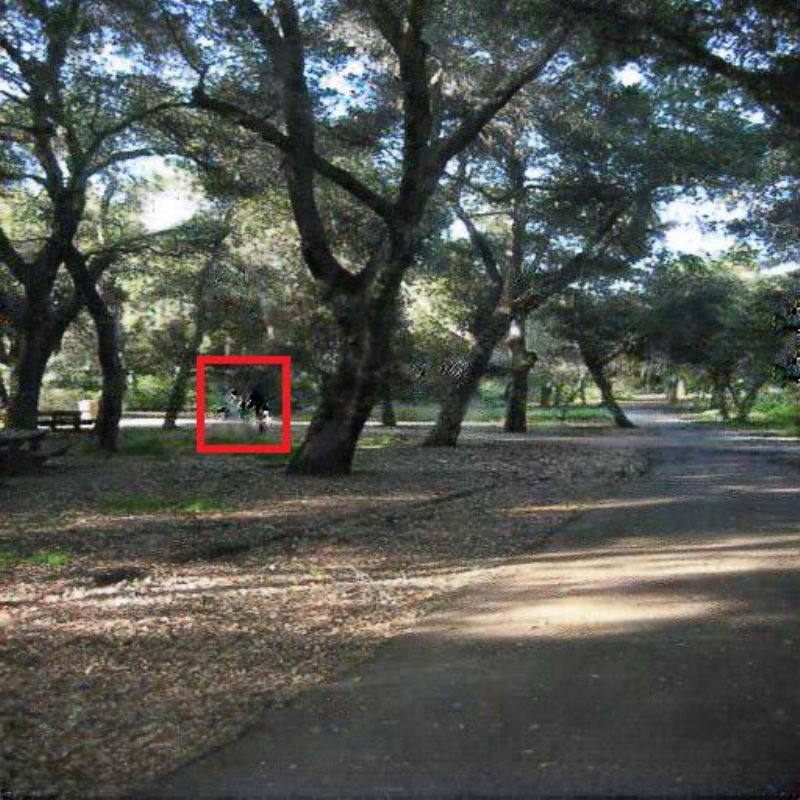} & \hspace{-4mm}
\includegraphics[width = 0.24\linewidth]{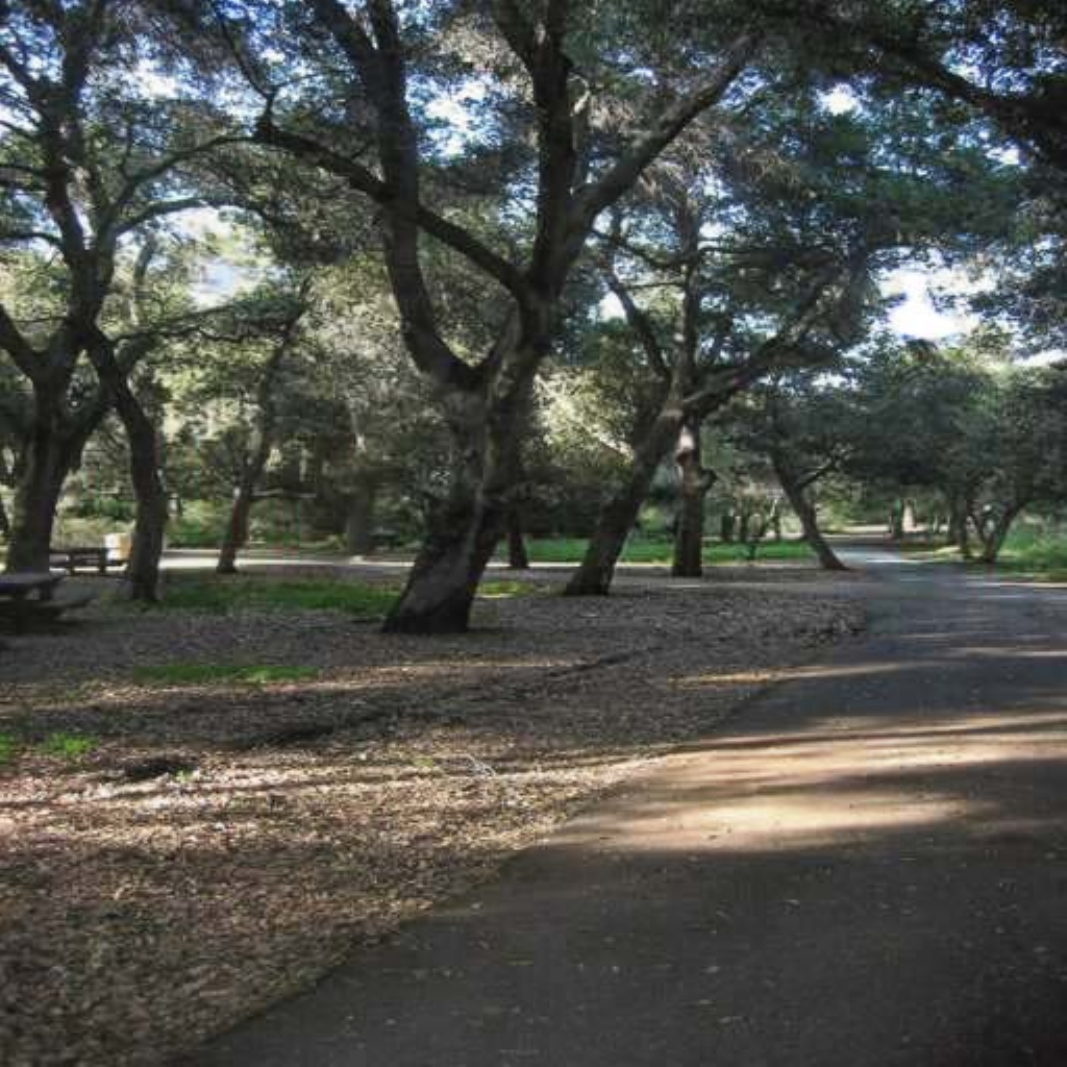}& \hspace{-4mm}
\includegraphics[width = 0.24\linewidth]{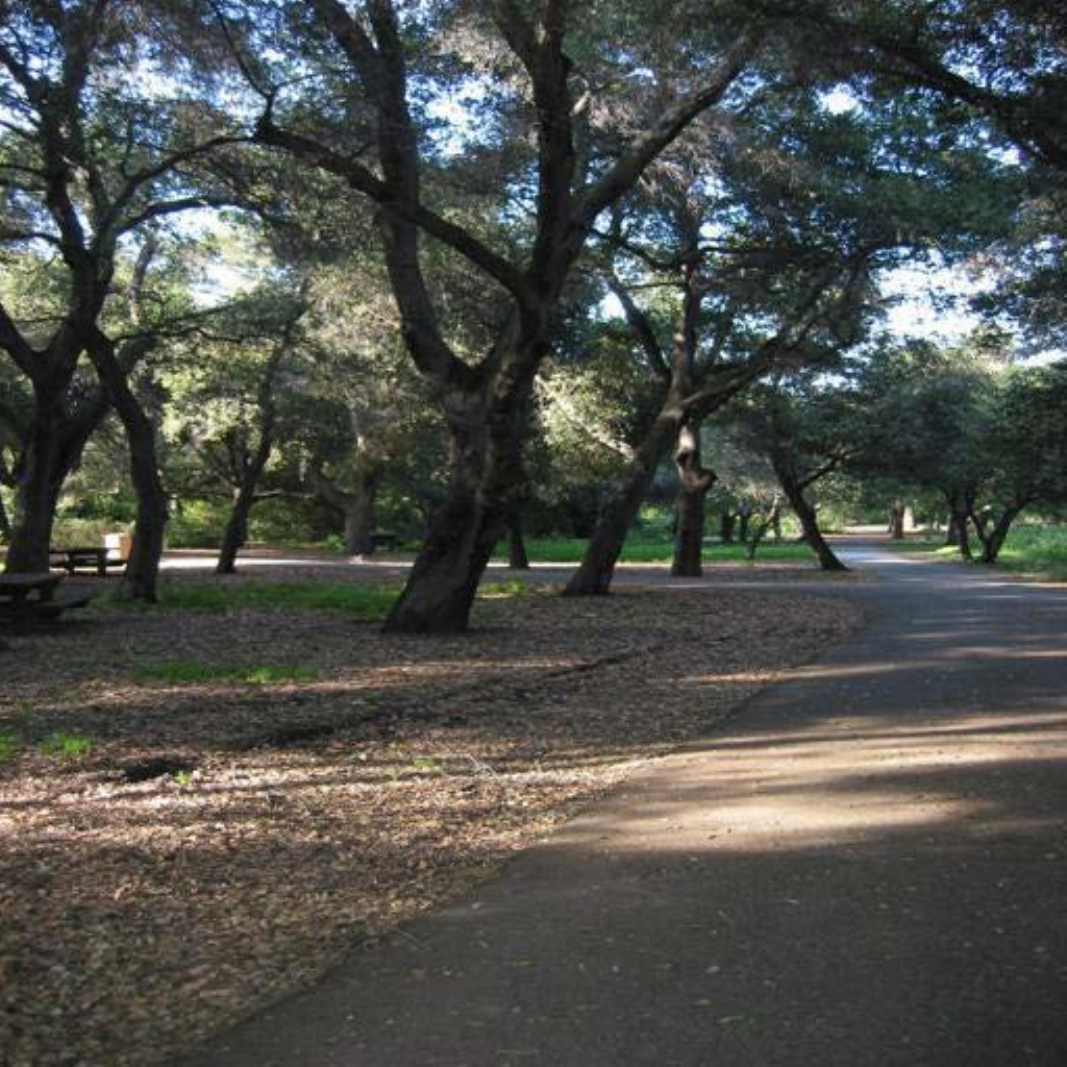}\\
 (e) Cai~\cite{DehazeNet/tip16} &\hspace{-4mm} (f) pix2pix~\cite{pixel/to/pixel} &\hspace{-4mm} (g) Ours &\hspace{-4mm} (h) GT\\
\end{tabular}
\caption{One synthetic hazy image from the proposed synthetic hazy dataset. The structures enclosed in the red box in (f) are not preserved well.
The proposed method does not need to estimate the transmission map and atmospheric light and generates a much clearer image which is visually close to the ground truth image (best viewed on high-resolution displays with zoom-in).
}
\label{fig: dehazing-synthetic}
\vspace{-1mm}
\end{figure*}

\vspace{-3mm}
\subsection{Image Dehazing}
\vspace{-1mm}
For image dehazing, we evaluate the proposed algorithm against the state-of-the-art methods based on
statistical priors~\cite{he/dark/channel/dehazing/cvpr09,non/local/dehazing,juewang/dehazing/eccv16} and deep models~\cite{dehaze/eccv16,DehazeNet/tip16,zhanghe/cvpr18/dehazing}.
As the methods based on neural networks~\cite{dehaze/eccv16,DehazeNet/tip16}
estimate transmission maps
and the training code is not available~\cite{DehazeNet/tip16},
we use the provided models for comparisons.
We also retrain pix2pix~\cite{pixel/to/pixel} and CycleGAN~\cite{CycleGAN2017} using the proposed training dataset for fair comparisons.
\vspace{-2mm}
{\flushleft \bf{Synthetic hazy images.}}
We first evaluate our method on the synthetic dataset as mentioned in Section~\ref{ssec: datasets}.
The test dataset contains 241 images including both indoor and outdoor scenes.
Table~\ref{tab: psnr-dehazing} shows that the proposed algorithm performs favorably against state-of-the-art methods in terms of PSNR and SSIM.

We show one example from the test dataset in Figure~\ref{fig: dehazing-synthetic}. The conventional methods~\cite{he/dark/channel/dehazing/cvpr09,non/local/dehazing} need to estimate both the transmission map and atmospheric light based on manually designed features.
The colors of the generated images of~\cite{he/dark/channel/dehazing/cvpr09} and~\cite{non/local/dehazing} are slightly different from those of the ground truth images due to inaccurate estimations of the transmission map and atmospheric light.
The CNN-based methods~\cite{dehaze/eccv16,DehazeNet/tip16} first estimate the transmission maps from hazy images and use the conventional schemes to restore clear images.
The restored images contain hazy residual as shown in Figure~\ref{fig: dehazing-synthetic}(d)-(e).
Although the GAN-based methods (e.g., pix2pix~\cite{pixel/to/pixel}) are able to generate realistic images, the structures are not preserved well.
In contrast, due to the use of the image degradation model constraint, the proposed algorithm generates a much clearer image with fine details, which is visually close to the ground truth image.

\begin{figure*}[!t]
\centering
\begin{tabular}{cccc}
\includegraphics[width = 0.24\linewidth]{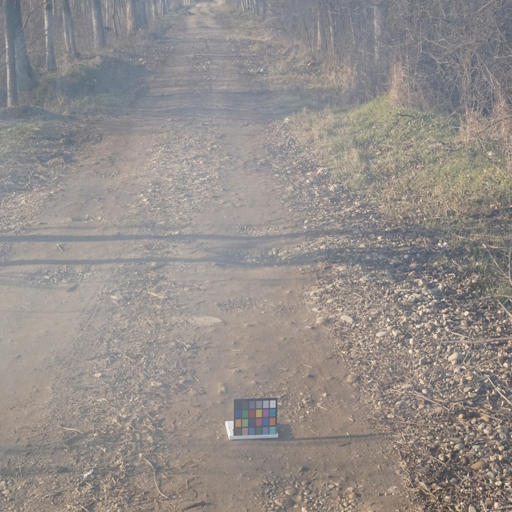}& \hspace{-4mm}
\includegraphics[width = 0.24\linewidth]{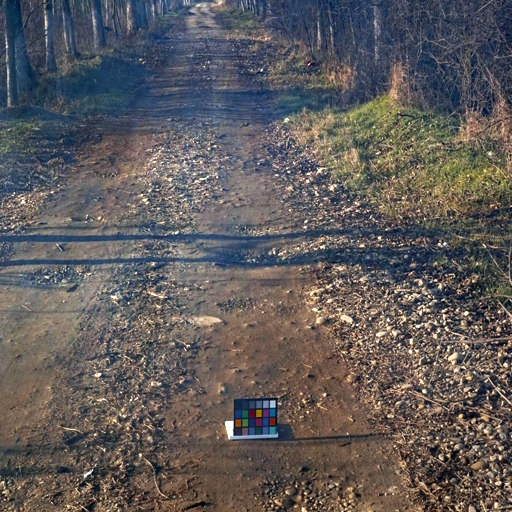}& \hspace{-4mm}
\includegraphics[width = 0.24\linewidth]{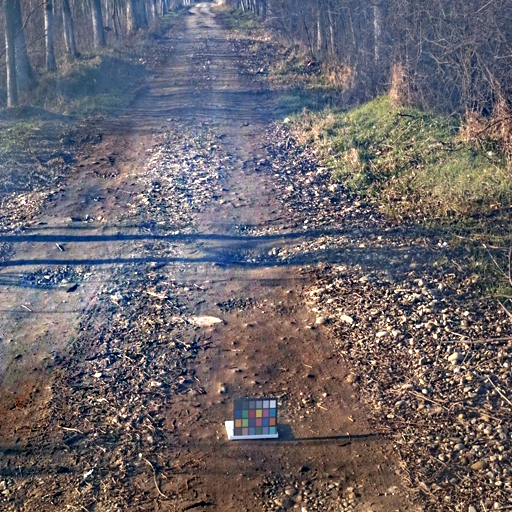}& \hspace{-4mm}
\includegraphics[width = 0.24\linewidth]{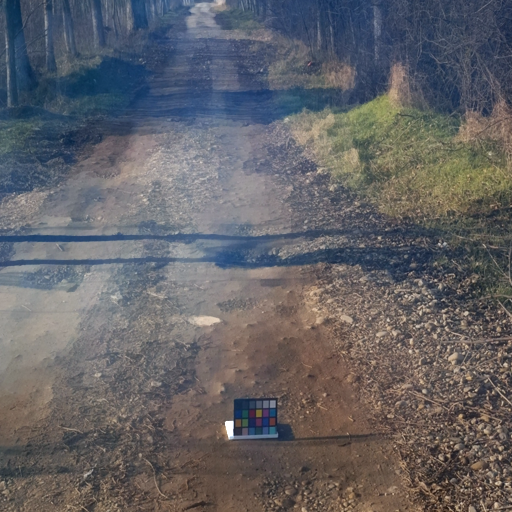} \\
(a) Input &\hspace{-4mm} (b) He~\cite{he/dark/channel/dehazing/cvpr09} &\hspace{-4mm} (c)  Berman~\cite{non/local/dehazing} &\hspace{-4mm} (d) Chen~\cite{juewang/dehazing/eccv16}\\
\includegraphics[width = 0.24\linewidth, height = 0.16\linewidth]{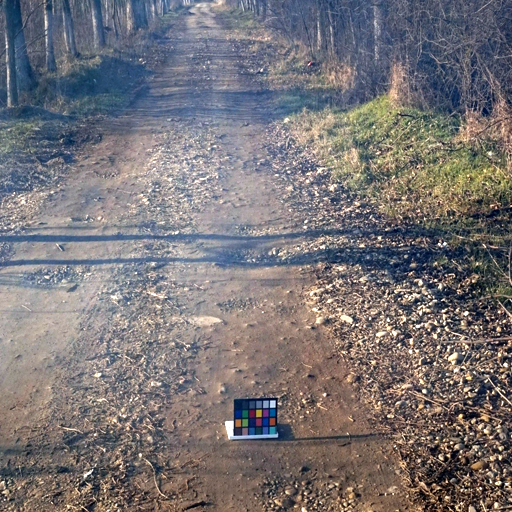}& \hspace{-4mm}
\includegraphics[width = 0.24\linewidth]{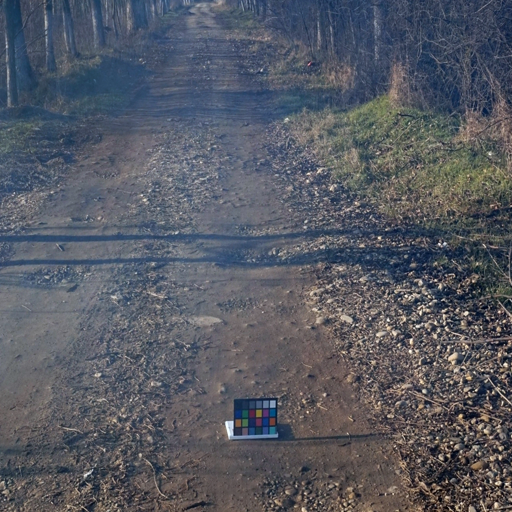}& \hspace{-4mm}
\includegraphics[width = 0.24\linewidth]{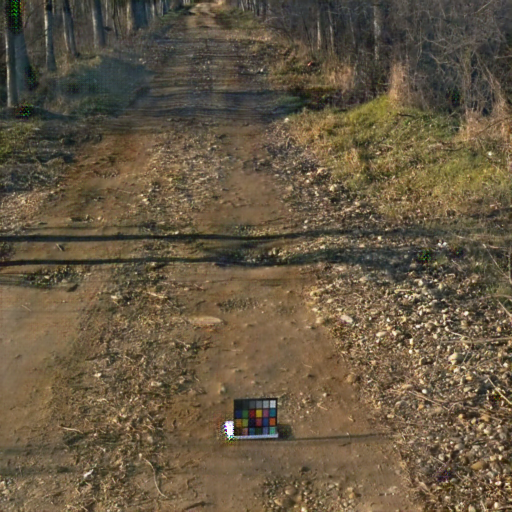}& \hspace{-4mm}
\includegraphics[width = 0.24\linewidth]{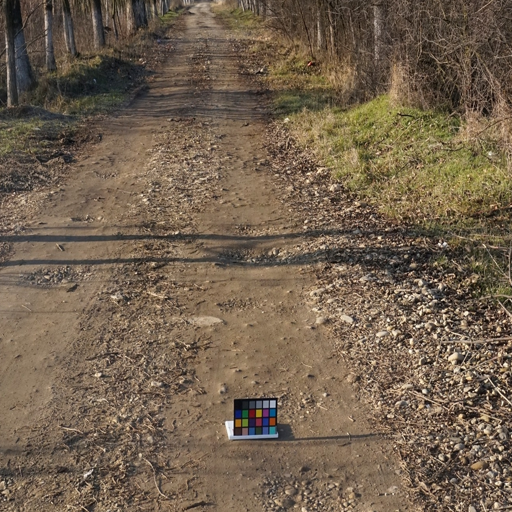}\\
 (e) Ren~\cite{dehaze/eccv16}  &\hspace{-4mm}  (f) Cai~\cite{DehazeNet/tip16} &\hspace{-4mm} (g) Ours &\hspace{-4mm}  (h) GT\\
\end{tabular}
\caption{A hazy image from the dataset~\cite{ntire18}. The proposed method generates a much clearer image which is visually close to the ground truth.
}
\label{fig: dehazing-ntire}
\vspace{-1mm}
\end{figure*}

We further evaluate the proposed algorithm on the outdoor dataset~\cite{ntire18} which contains 45 images.
Although the hazy images are captured in hazy environments, we refer to this dataset as a synthetic dataset because the haze is manmade.
As the resolution of the original test images is too large, we resize each test image to the size of $512 \times 512$ pixels for comparisons.
Table~\ref{tab: psnr-dehazing-ntire18} and Figure~\ref{fig: dehazing-ntire} show that the proposed algorithm performs favorably against state-of-the-art dehazing methods.

\begin{table*}[!t]
\caption{\label{tab: psnr-dehazing-ntire18} Quantitative evaluations with the state-of-the-art methods on the outdoor image dehazing dataset~\cite{ntire18}. The results of compared methods are generated by the provided implementations or models.
}
\vspace{-3mm}
\centering
\begin{tabular}{cccccccccc}
\toprule
Methods      & Input       & He~\cite{he/dark/channel/dehazing/cvpr09} & Berman~\cite{non/local/dehazing} & Chen~\cite{juewang/dehazing/eccv16} & Meng~\cite{meng/dehaze/iccv13} &Ren~\cite{dehaze/eccv16} & Cai~\cite{DehazeNet/tip16}& Zhang~\cite{zhanghe/cvpr18/dehazing}& Ours \\
\midrule
PSNR         & 13.89          & 16.60                                        & 15.76                                & 15.92  & 16.04           & 17.44&  15.66     &      13.08        &\bf{17.76}\\
SSIM         & 0.6296         & 0.6915                                        & 0.7483                                & 0.6322  & 0.7374        &0.7551   &   0.6744   &      0.7301       &\bf{0.7795} \\
\bottomrule
\end{tabular}
\vspace{-2mm}
\end{table*}

\begin{figure*}[!t]
\centering
\begin{tabular}{cccc}
\includegraphics[width = 0.24\linewidth]{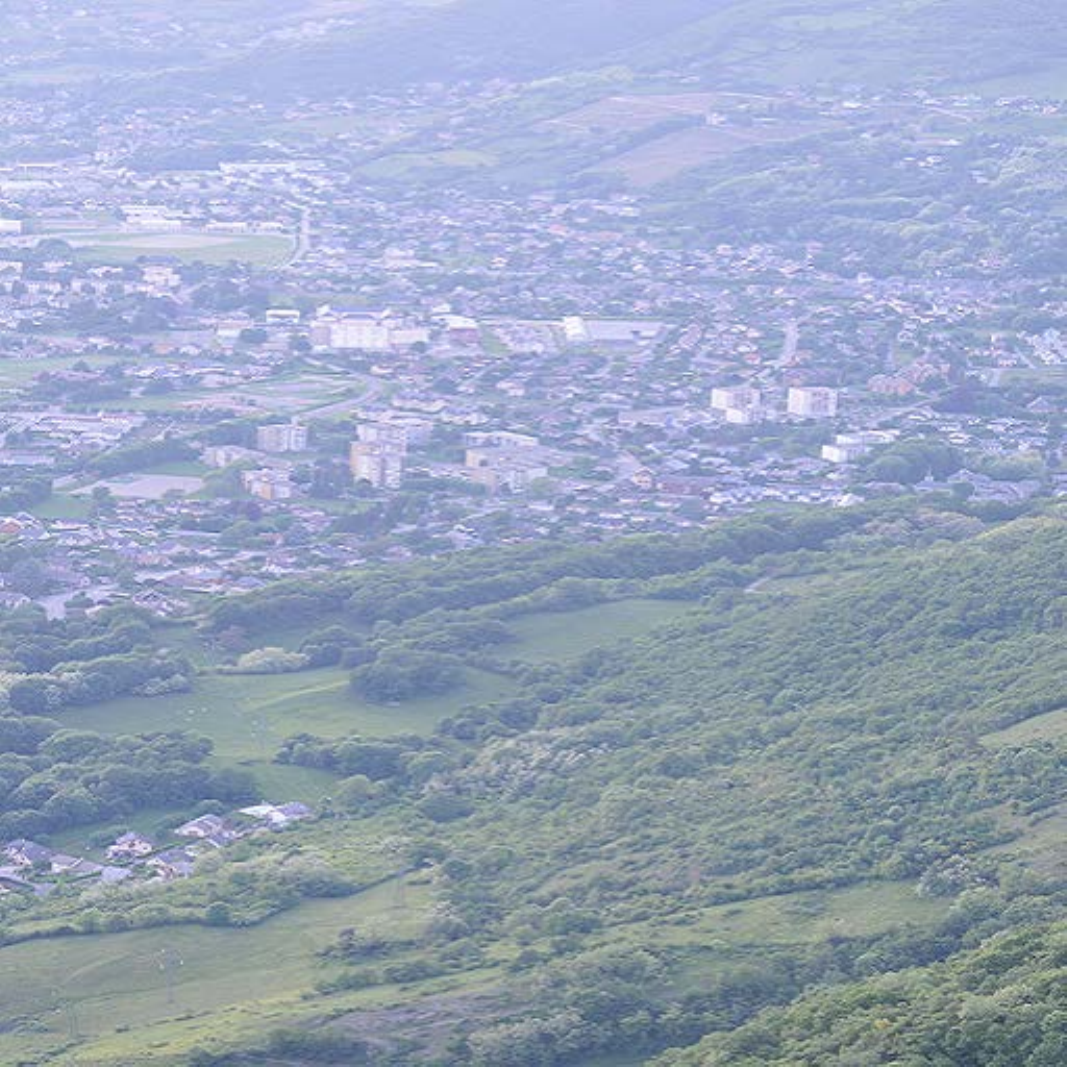}& \hspace{-4mm}
\includegraphics[width = 0.24\linewidth]{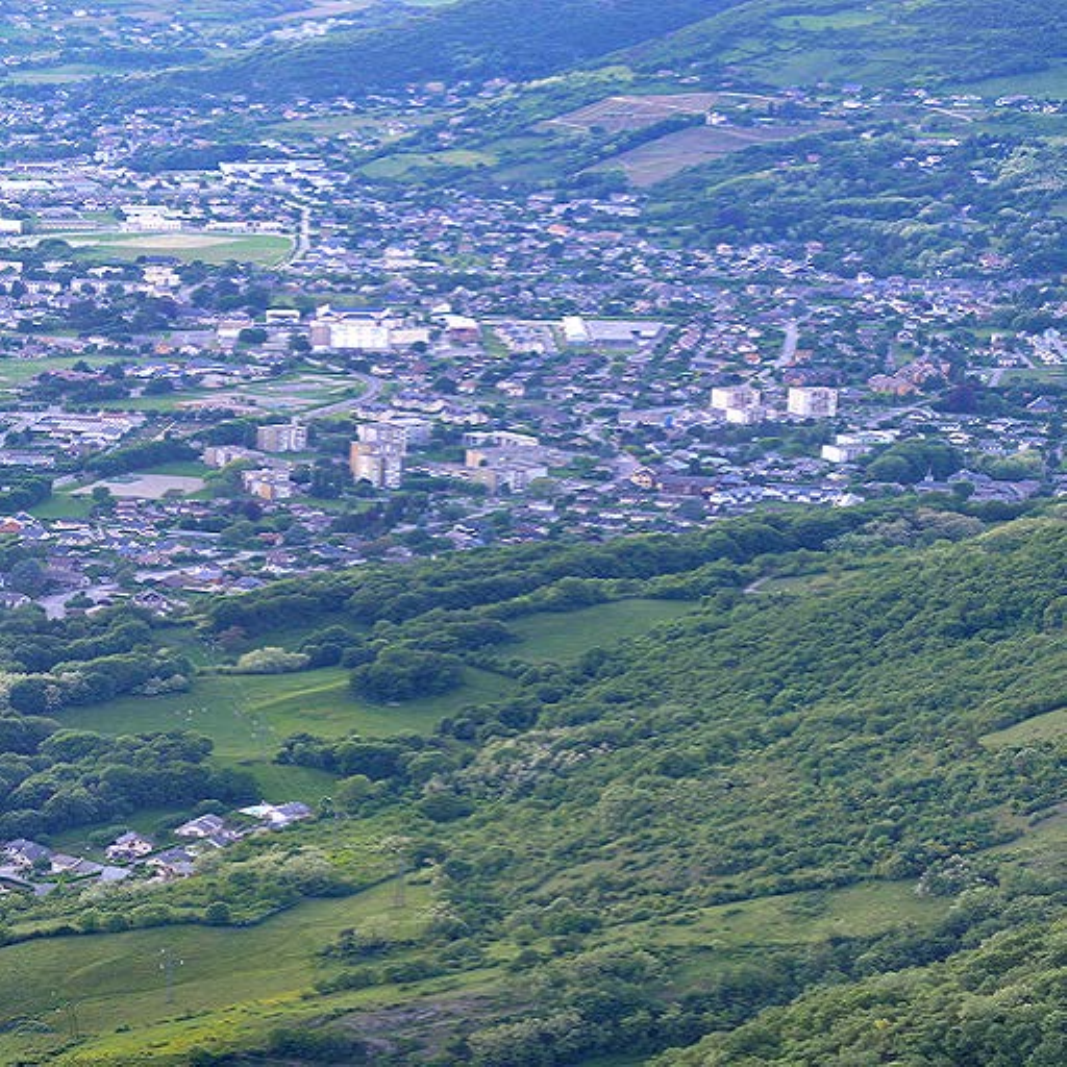}& \hspace{-4mm}
\includegraphics[width = 0.24\linewidth]{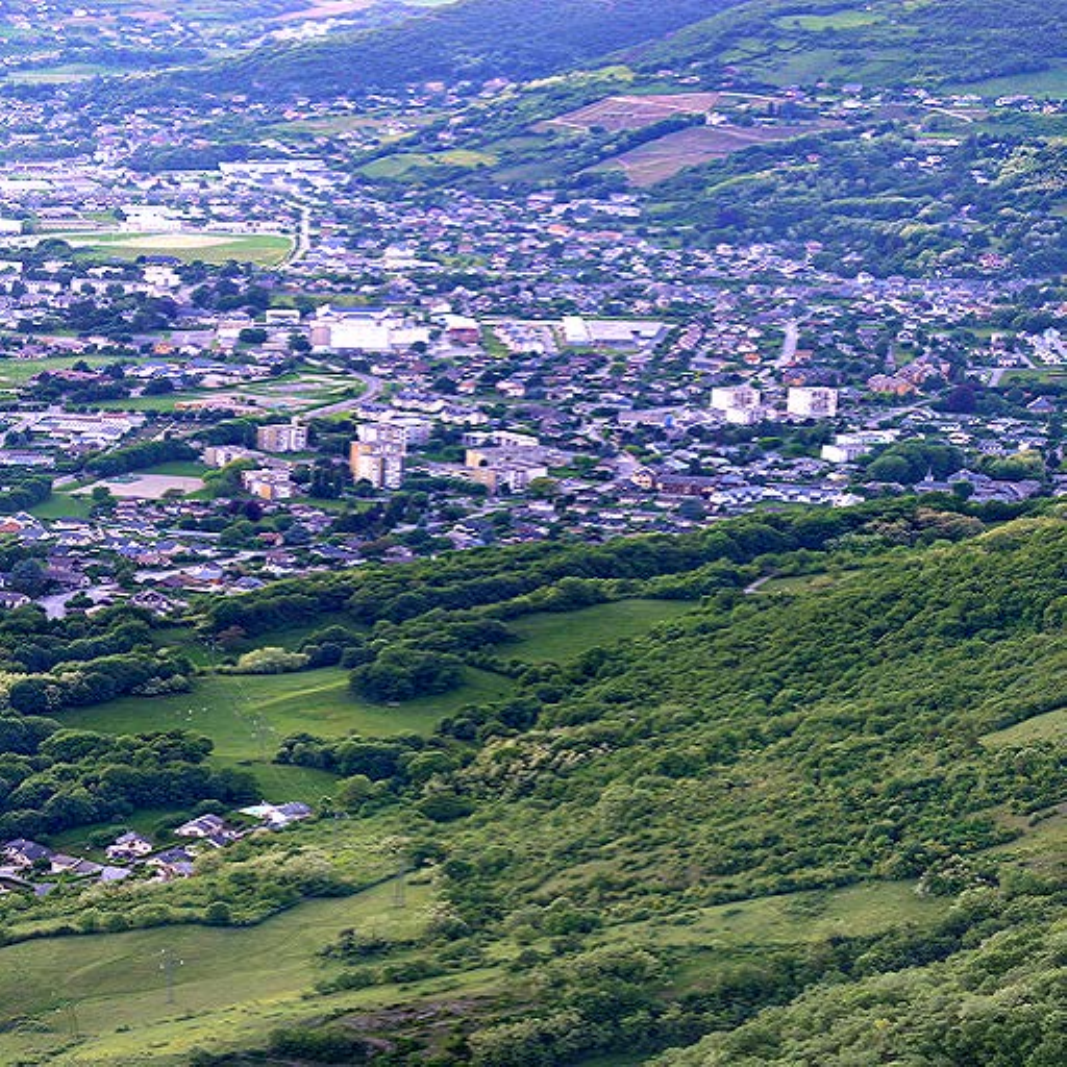}& \hspace{-4mm}
\includegraphics[width = 0.24\linewidth]{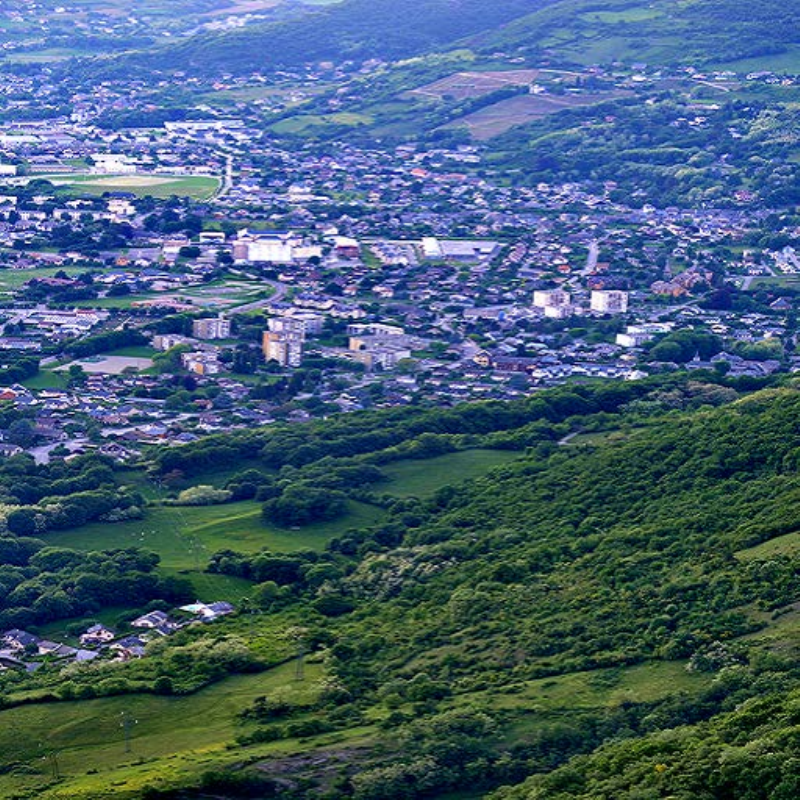} \\
(a) Input &\hspace{-4mm} (b) He~\cite{he/dark/channel/dehazing/cvpr09} &\hspace{-4mm} (c)  Berman~\cite{non/local/dehazing} &\hspace{-4mm} (d) Ren~\cite{dehaze/eccv16}\\
\includegraphics[width = 0.24\linewidth]{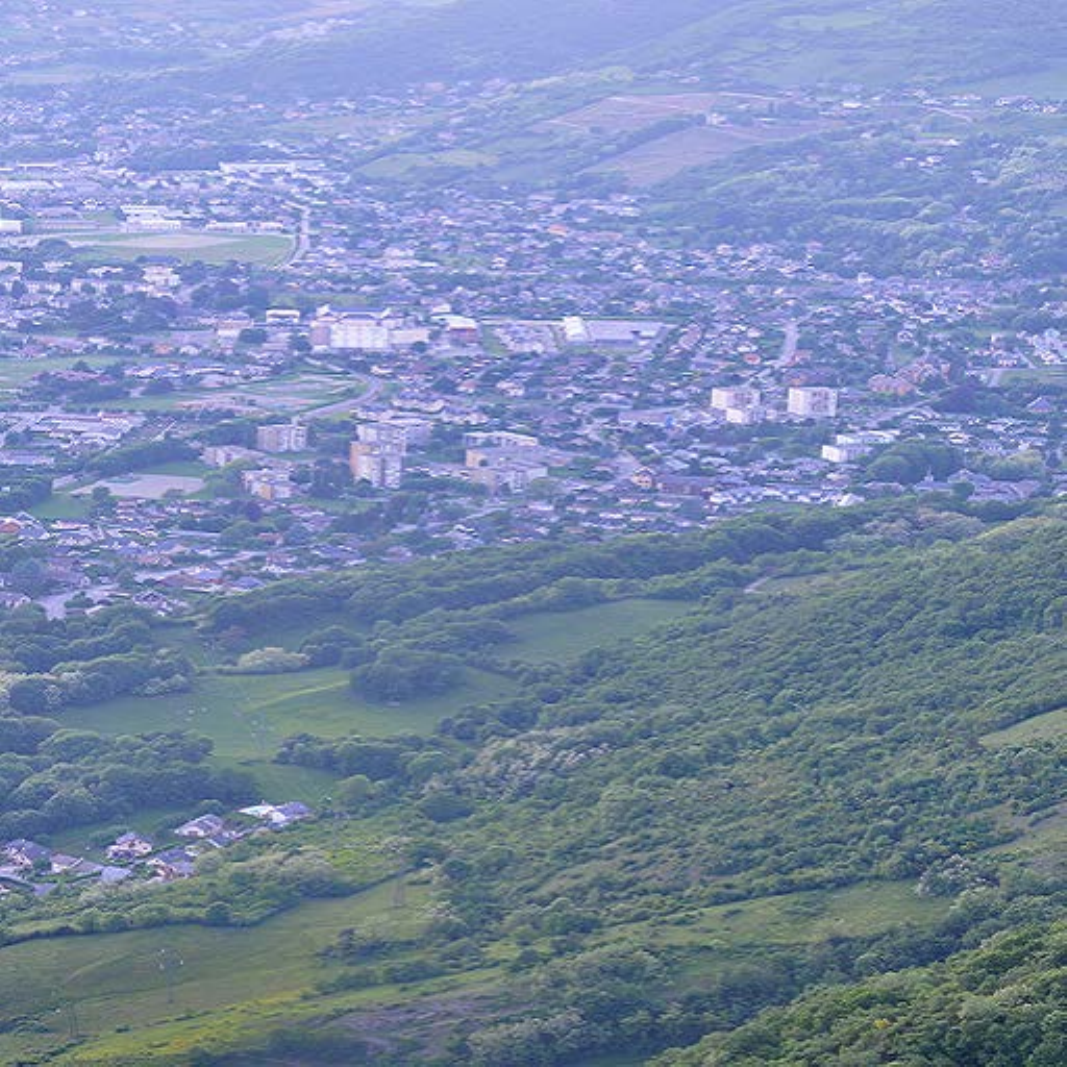}& \hspace{-4mm}
\includegraphics[width = 0.24\linewidth]{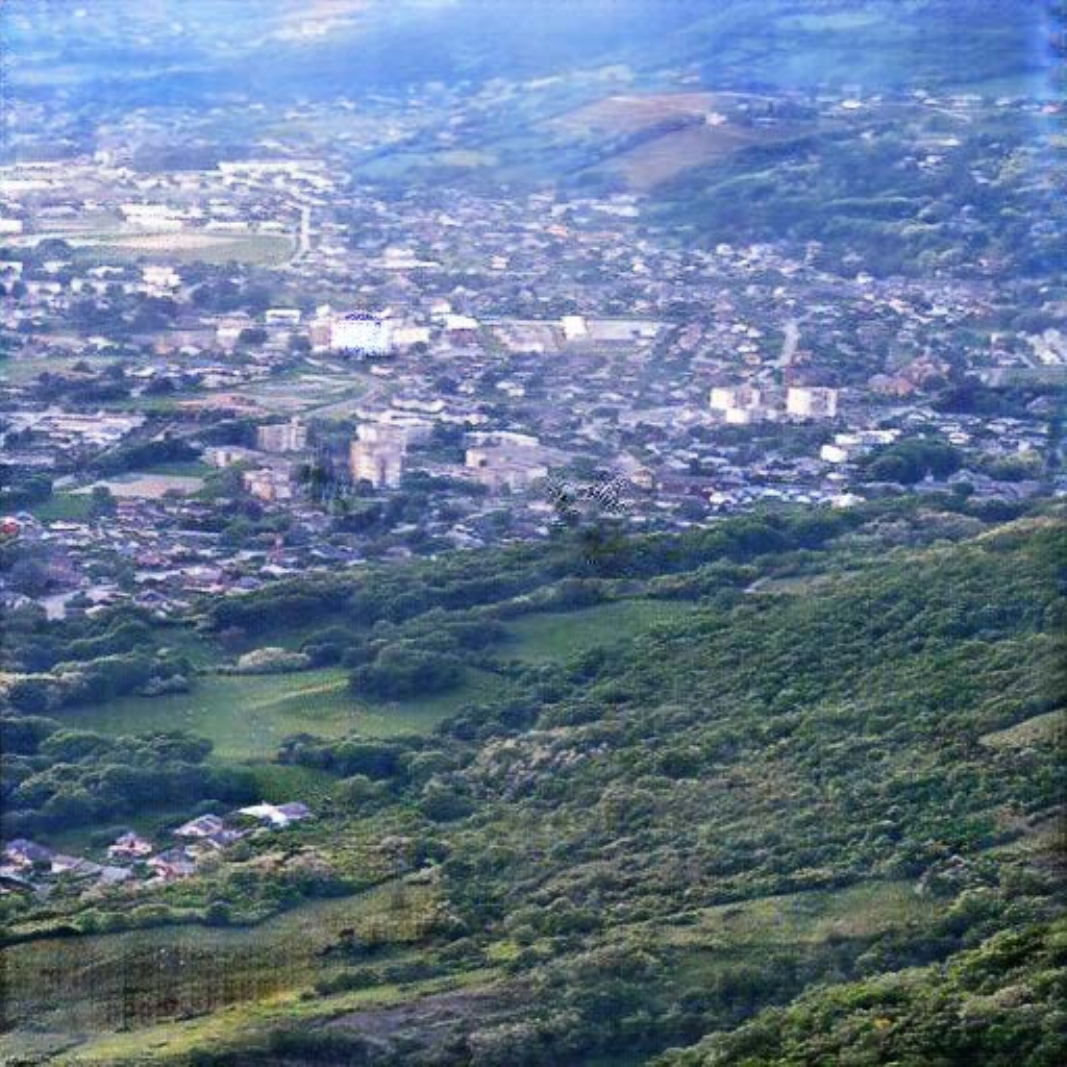}& \hspace{-4mm}
\includegraphics[width = 0.24\linewidth]{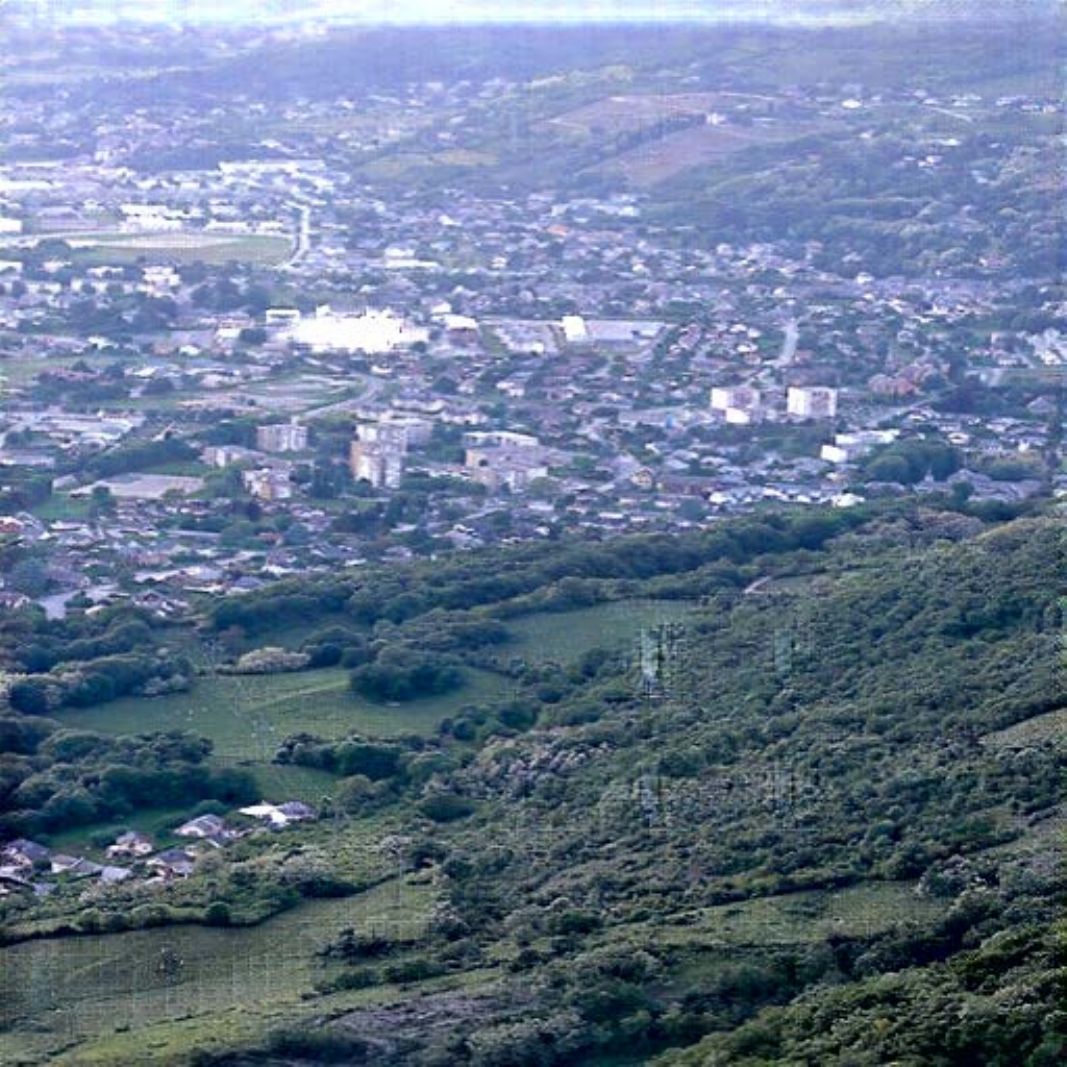}& \hspace{-4mm}
\includegraphics[width = 0.24\linewidth]{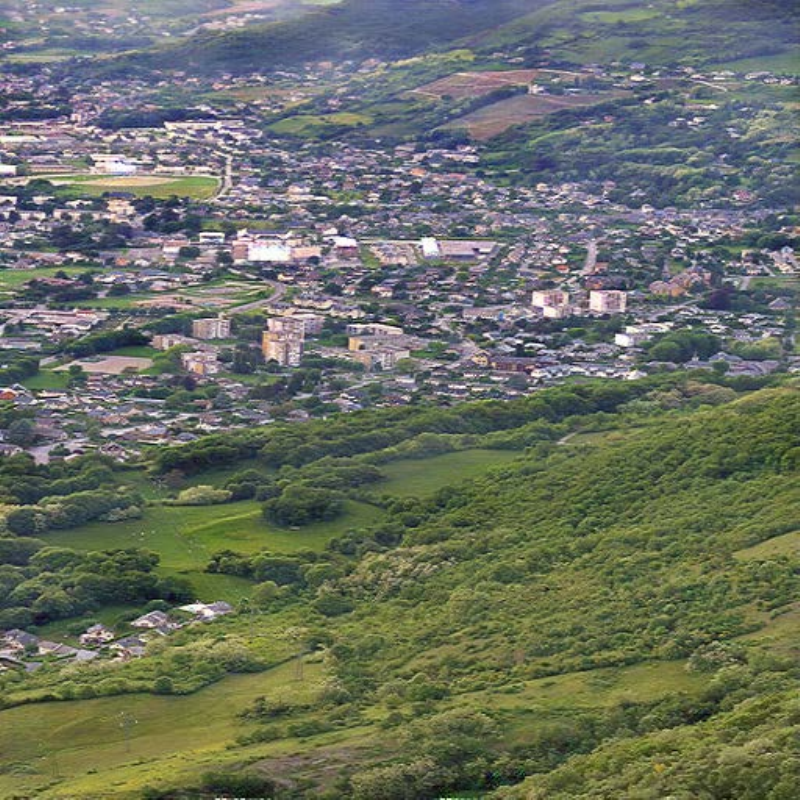}\\
 (e) Cai~\cite{DehazeNet/tip16} &\hspace{-4mm}  (f) pix2pix~\cite{pixel/to/pixel} &\hspace{-4mm} (g) PCycleGAN~\cite{CycleGAN2017} &\hspace{-4mm}  (h) Ours\\
\end{tabular}
\caption{Results on a real hazy image. The proposed method generates a clearer image.
}
\label{fig: dehazing-real}
\vspace{-1mm}
\end{figure*}

\vspace{-2mm}
{\flushleft \bf{Real hazy images.}}
We evaluate the proposed algorithm on real hazy images
and show some results against the state-of-the-art methods in Figure~\ref{fig: dehazing-real}.
The colors of the restored image by the proposed method are more vivid than the other algorithms.

\vspace{-3mm}
\subsection{Image Super-resolution}
\vspace{-1mm}
\label{ssec: Image Super-resolution}
The proposed algorithm can be applied to image super-resolution by replacing the physics model~\eqref{eq: mapping-function-low-level} with the image formulation of super-resolution as stated in Section~\ref{sec: proposed-algorithm}.
We randomly choose 50,000 images from the COCO dataset~\cite{coco} to train the proposed algorithm and evaluate it on the images from the ``Set5" against the state-of-the-art algorithms including SRCNN~\cite{SRCNN/pami16}, ESPCN~\cite{ESPCN}, VDSR~\cite{VDSR/cvpr16}, SRGAN~\cite{SRGAN}, and EDSR~\cite{Yang_2017_CVPR}.
We fine-tune the methods based on  deep neural networks using the proposed training dataset and choose the best models for fair comparisons.
Quantitative evaluation results are shown in Table~\ref{tab: psnr-sr-ntire18}.
Although the proposed algorithm is not designed for super-resolution, it achieves competitive results compared to the state-of-the-art methods.
\begin{table*}[!t]
\caption{\label{tab: psnr-sr-ntire18} Quantitative evaluations with the state-of-the-art methods on the image super-resolution ($\times4$) problem.
}
\vspace{-3mm}
\centering
\begin{tabular}{ccccccccc}
\toprule
Methods      & Bicubic       & SRCNN~\cite{SRCNN/pami16} & ESPCN~\cite{ESPCN}  & VDSR~\cite{VDSR/cvpr16} & SRGAN~\cite{SRGAN}   & EDSR~\cite{edsr}      & Ours \\
\midrule
PSNR         & 28.42            & 30.50                         & 30.27             & 31.41                      & 32.05                   & \bf{32.46}                          & 30.03   \\
SSIM         & 0.8104          & 0.8629                         & 0.8540             & 0.8840                     & 0.8910                  & 0.8968                         & \bf{0.9030}   \\
\bottomrule
\end{tabular}
\vspace{-2mm}
\end{table*}

\vspace{-3mm}
\subsection{Image Deraining}
\label{ssec: Image Deraining}
\vspace{-1mm}
We further apply the proposed algorithm to image deraining problem which aims to remove rainy streaks or dirties from the input images.
The image deraining model is based on a linear  superimposition of clear image and rainy streak.
We evaluate the proposed algorithm against the conventional image prior based method~\cite{liyu/derain/cvpr16} and state-of-the-art deep learning based methods~\cite{derain/gan,Yang_2017_CVPR,fu/derain/tip17} using the test dataset by~\cite{derain/gan} for comparisons.
We also retrain pix2pix~\cite{pixel/to/pixel}, CycleGAN~\cite{CycleGAN2017}, and PCycleGAN using the same training dataset~\cite{derain/gan} for fair comparisons.

\begin{figure*}[!t]
\centering
\begin{tabular}{cccc}
\includegraphics[width = 0.24\linewidth, height = 0.17\linewidth]{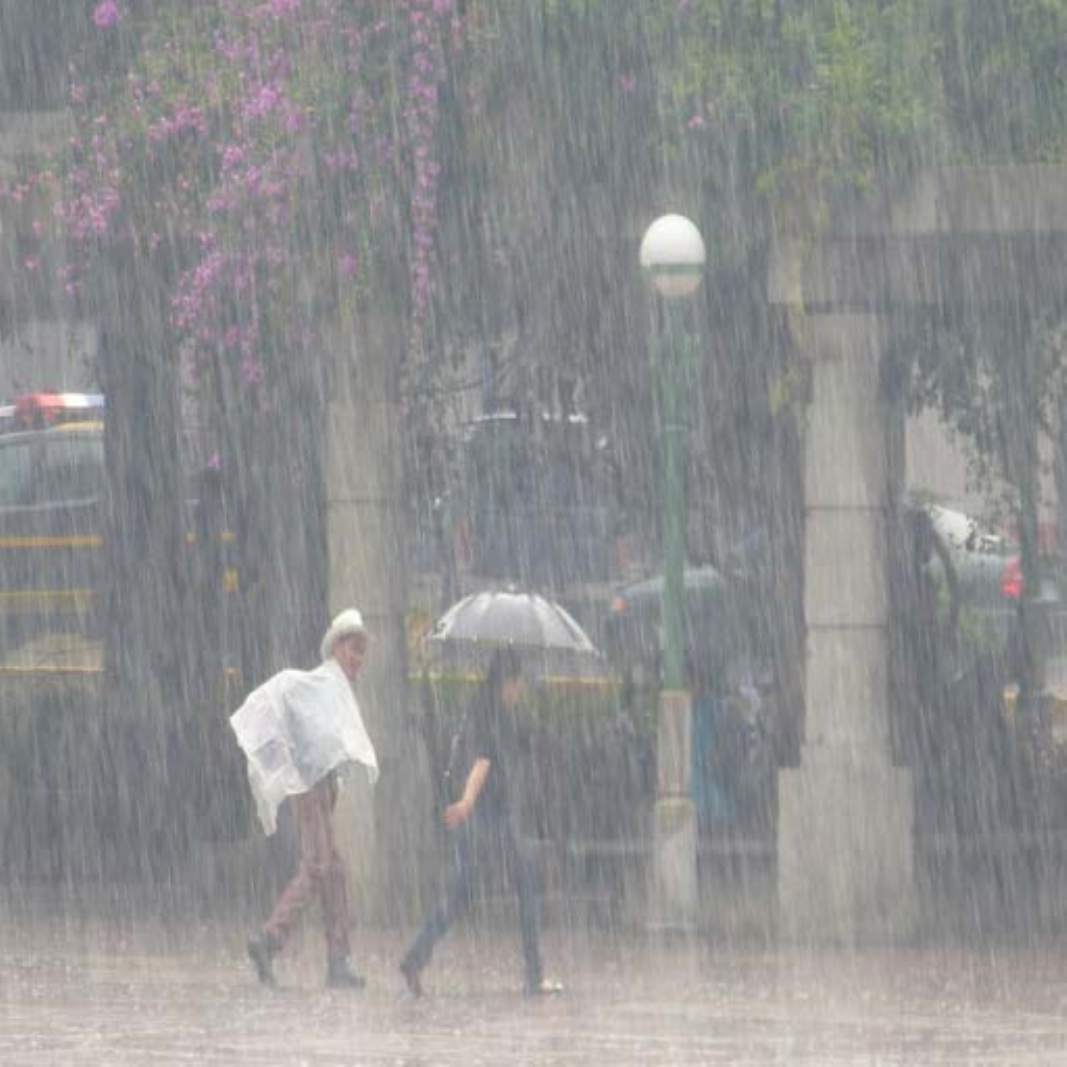}& \hspace{-4mm}
\includegraphics[width = 0.24\linewidth, height = 0.17\linewidth]{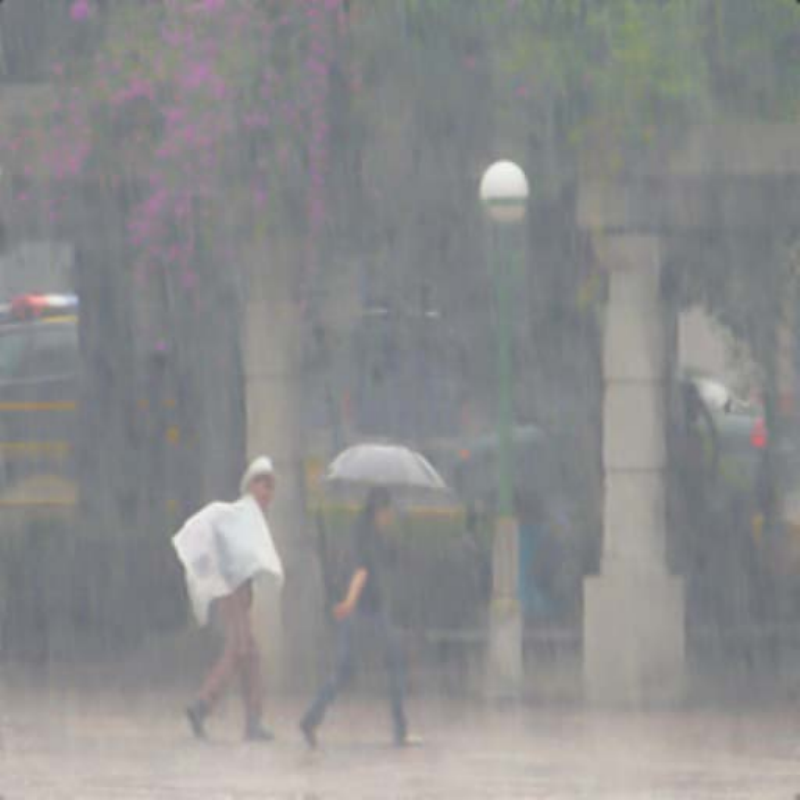}& \hspace{-4mm}
\includegraphics[width = 0.24\linewidth, height = 0.17\linewidth]{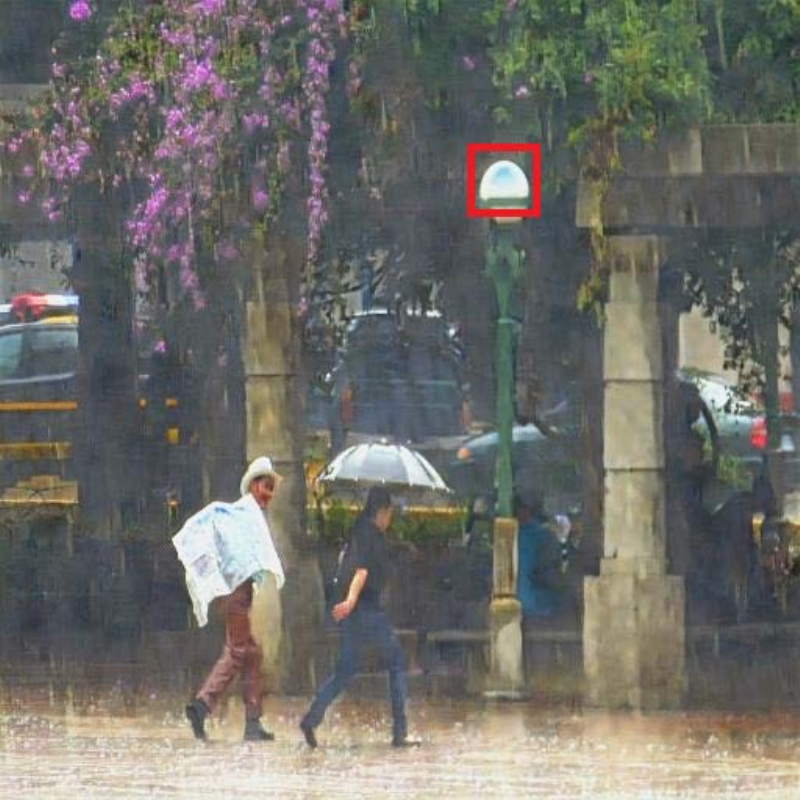} & \hspace{-4mm}
\includegraphics[width = 0.24\linewidth, height = 0.17\linewidth]{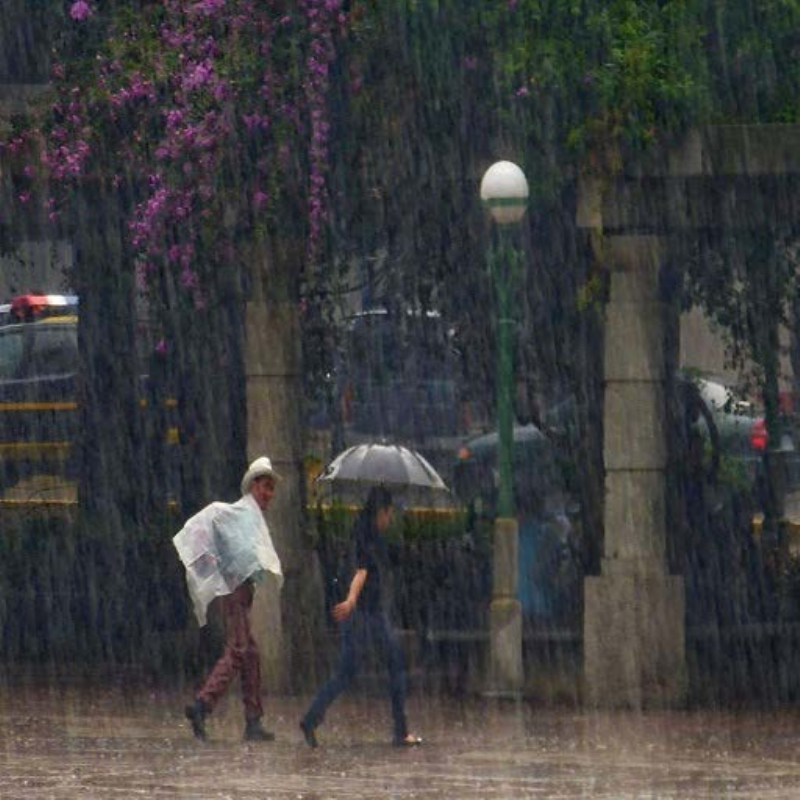} \\
(a) Input & \hspace{-4mm} (b) Li~\cite{liyu/derain/cvpr16} &\hspace{-4mm} (c) Zhang~\cite{derain/gan} &\hspace{-4mm} (d) Fu~\cite{fu/derain/tip17} \\
\includegraphics[width = 0.24\linewidth, height = 0.17\linewidth]{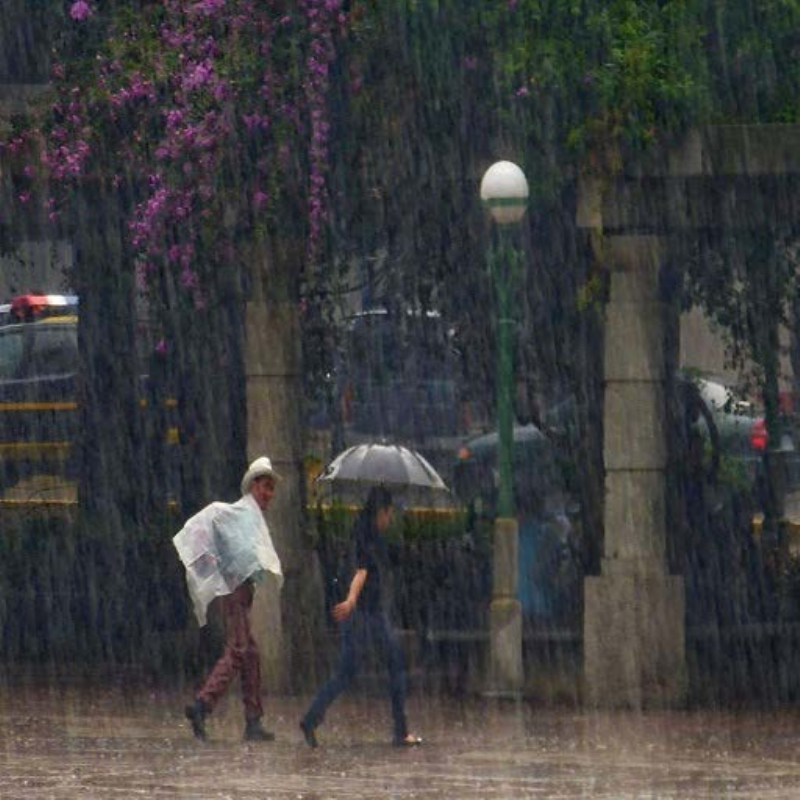}& \hspace{-4mm}
\includegraphics[width = 0.24\linewidth, height = 0.17\linewidth]{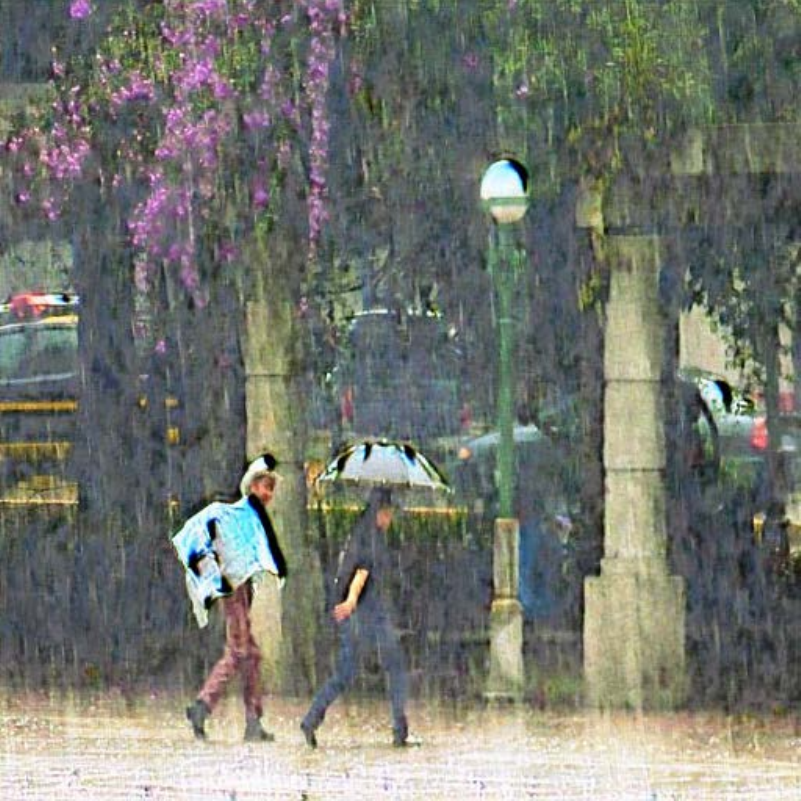} & \hspace{-4mm}
\includegraphics[width = 0.24\linewidth, height = 0.17\linewidth]{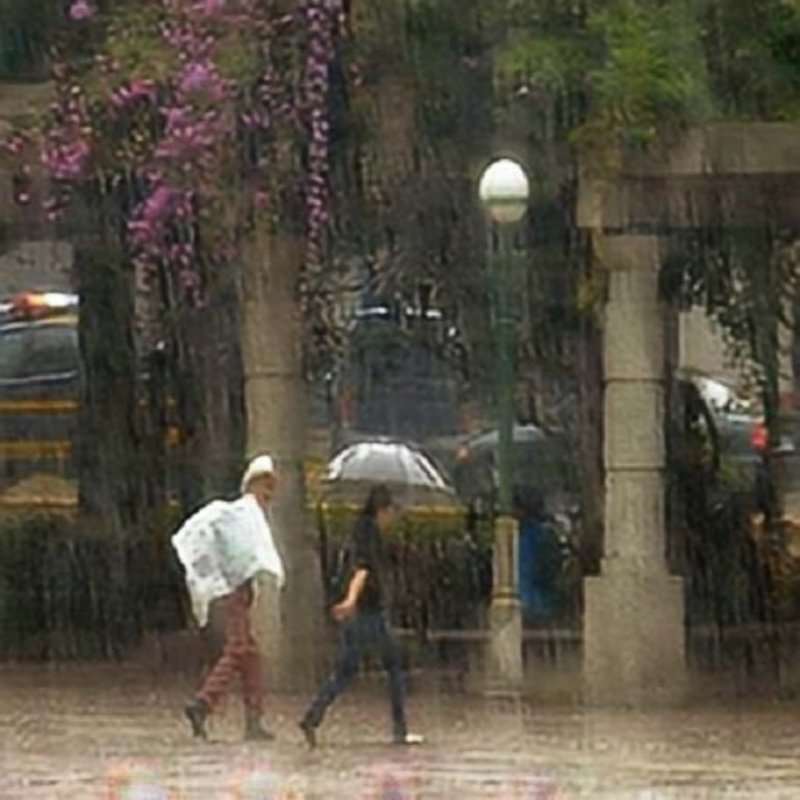}& \hspace{-4mm}
\includegraphics[width = 0.24\linewidth, height = 0.17\linewidth]{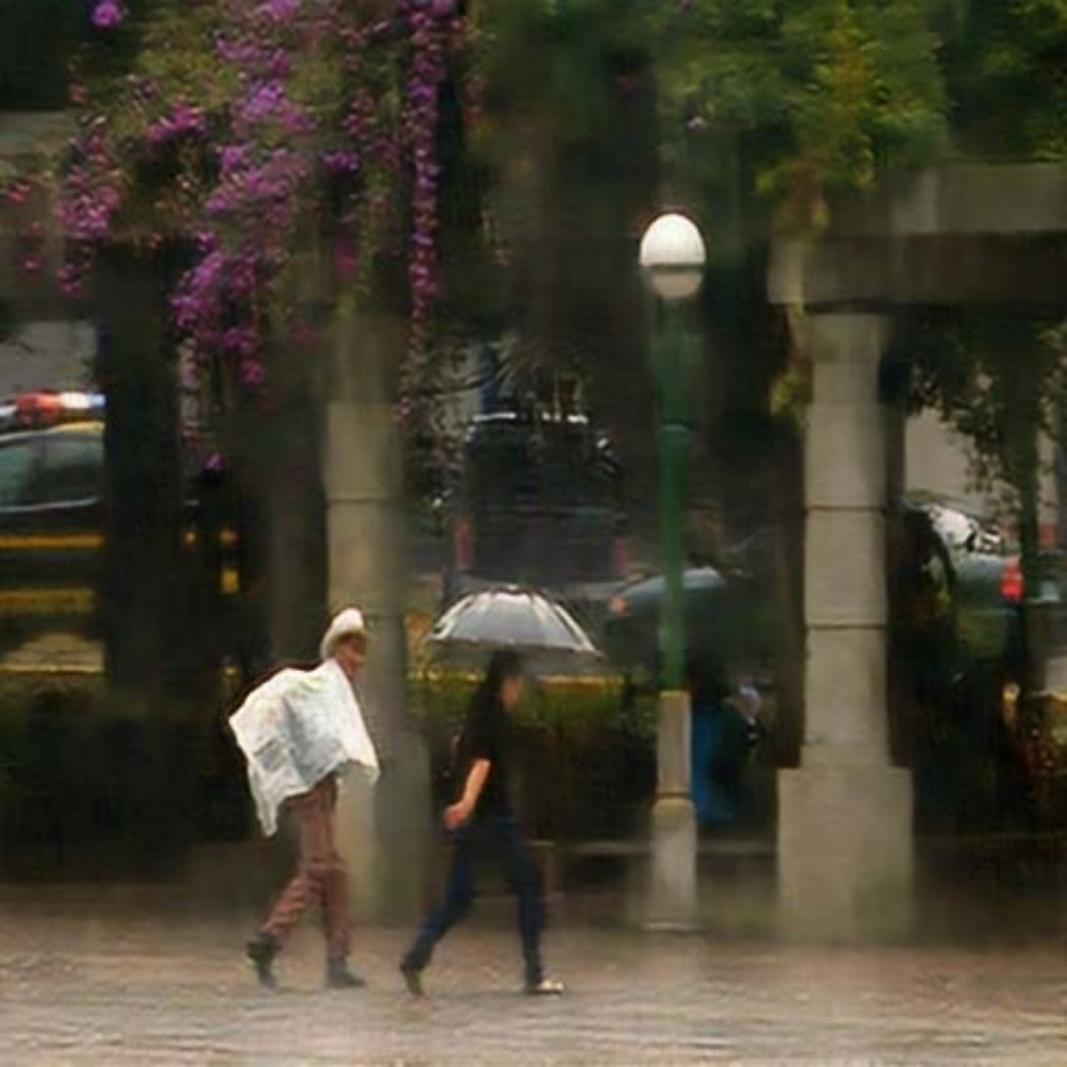} \\
(e) Yang~\cite{Yang_2017_CVPR} &\hspace{-4mm} (f) pix2pix~\cite{pixel/to/pixel} &\hspace{-4mm} (g) PCycleGAN~\cite{CycleGAN2017} &\hspace{-4mm} (h) Ours\\
\end{tabular}
\vspace{-2mm}
\caption{Results on a real rainy image.
Due to the heavy rain, both the CNN-based~\cite{fu/derain/tip17,Yang_2017_CVPR} and GAN-based~\cite{derain/gan,pixel/to/pixel,CycleGAN2017} methods
are not able to generate clear images.
The part enclosed in the red box in (c) contains significant color distortions (best viewed on high-resolution displays with zoom-in).
}
\label{fig: deraining-real}
\end{figure*}

Figure~\ref{fig: deraining-real} shows a real rainy image.
The method~\cite{liyu/derain/cvpr16} based on handcrafted image priors does not remove rainy streaks well.
We note that the image deraining algorithm~\cite{derain/gan} is based on a GAN and a perceptual loss function~\cite{Perceptual/Loss}.
However, there exist significant color distortions in the restored image (e.g., the part in the red box in Figure~\ref{fig: deraining-real}(c)).
As discussed in Section~\ref{sec: introduction}, this is mainly because the algorithm~\cite{derain/gan} does not consider the image degradation model constraint.
The scheme by Fu et al.~\cite{fu/derain/tip17} decomposes the input image into a detail layer and a base layer, where the detail layer is estimated by a CNN.
However, this method is less effective for the scenes with heavy rain.
The method by Yang et al.~\cite{Yang_2017_CVPR} based on a multi-task network does not remove the rainy steaks well as shown in Figure~\ref{fig: deraining-real}(e).
The pix2pix~\cite{pixel/to/pixel} and PCycleGAN algorithms do not generate clear images.
In contrast, the proposed method is able to remove rainy streaks and generates a much clearer image.

We further valuate the proposed method on other related low-level vision problems and present more experimental results in the supplementary material.

\vspace{-3mm}
\section{Analysis and Discussion}
\label{sec: Analysis and Discussion}
\vspace{-1mm}
In this section, we analyze the performance of the proposed algorithm against the state-of-the-art methods.

\begin{figure*}[!t]
\footnotesize
\centering
\begin{tabular}{cccc}
\includegraphics[width = 0.22\linewidth]{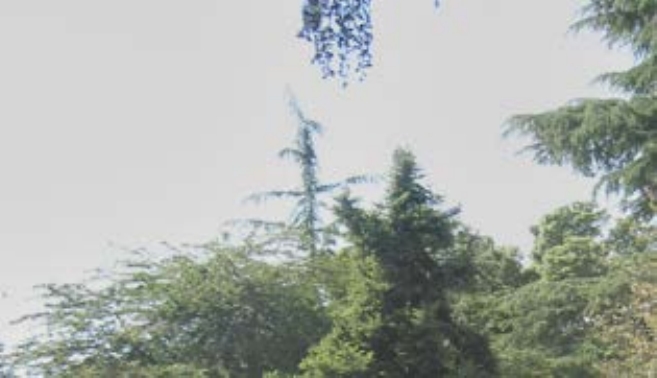}& \hspace{-4mm}
\includegraphics[width = 0.22\linewidth]{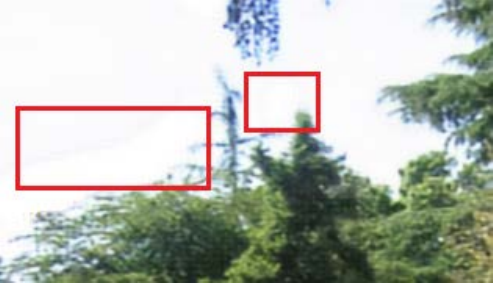}& \hspace{-4mm}
\includegraphics[width = 0.22\linewidth]{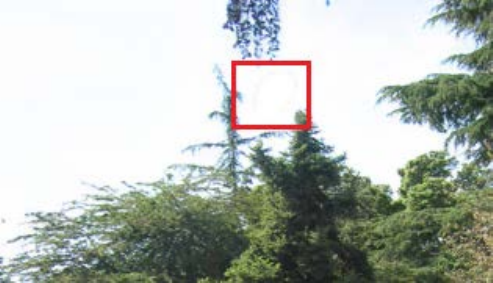}& \hspace{-4mm}
\includegraphics[width = 0.22\linewidth]{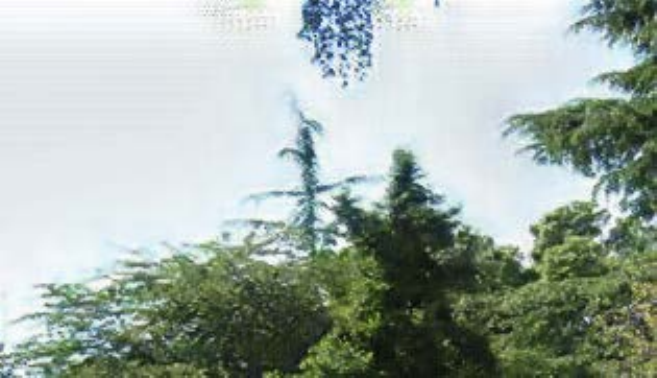}\\
(a) Input &\hspace{-4mm} (b) BaseGAN &\hspace{-4mm}  (c) w/o~\eqref{eq: pixel-wise-loss-co} &\hspace{-4mm} (d) w/o~\eqref{eq: pixel-wise-loss}\\
\includegraphics[width = 0.22\linewidth]{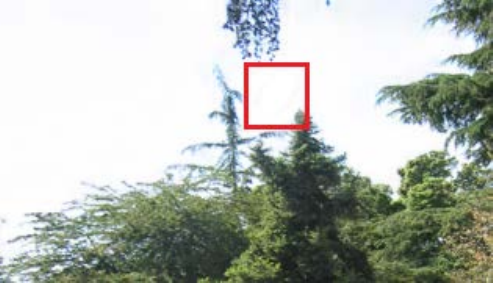} & \hspace{-4mm}
\includegraphics[width = 0.22\linewidth]{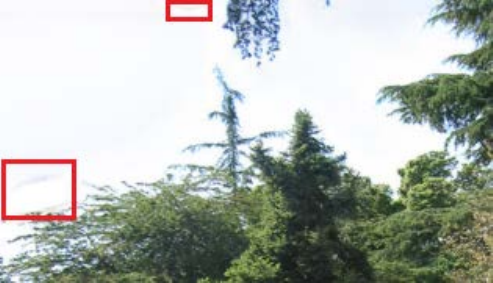}& \hspace{-4mm}
\includegraphics[width = 0.22\linewidth]{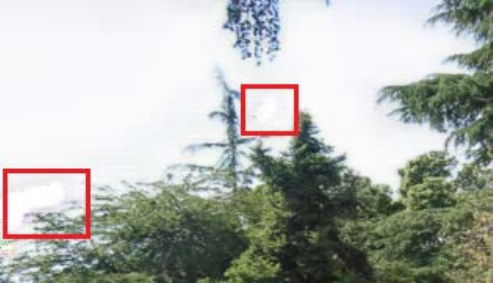}& \hspace{-4mm}
\includegraphics[width = 0.22\linewidth]{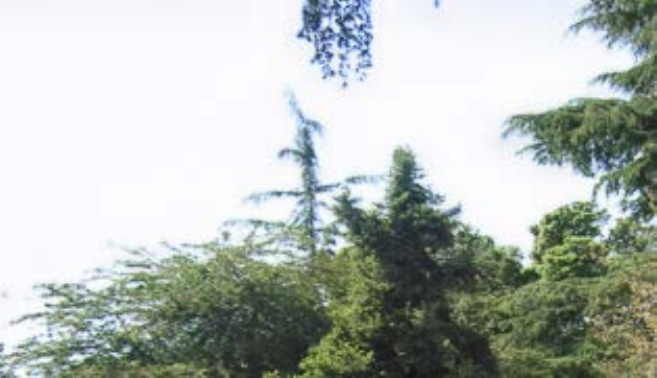}\\
(e) w/o~\eqref{eq: pixel-wise-loss-co-2}  &\hspace{-4mm} (f) w/o~\eqref{eq: pixel-wise-loss-co} \&~\eqref{eq: pixel-wise-loss-co-2} &\hspace{-4mm}  (g) PCycleGAN w/\eqref{eq: pixel-wise-loss-co}-\eqref{eq: pixel-wise-loss-co-2} &\hspace{-4mm}  (h) Ours\\
\end{tabular}
\vspace{-1mm}
\caption{
Visual comparison results of the proposed method with different loss functions on image dehazing.
The structures of the restored images (e.g., the sky and the parts enclosed in the red boxes) by the baseline methods are not estimated well.
}
\label{fig: effect-loss-function}
\vspace{-2mm}
\end{figure*}

\vspace{-2mm}
{\flushleft {\bf Effect of the image degradation model constraint.}}
Our method without the image degradation model constraint reduces to the GAN model~\cite{GAN} with the loss function~\eqref{eq: pixel-wise-loss}.
To analyze the effect of this constraint with fair comparisons, we remove the physics model and corresponding loss functions (BaseGAN for short) in our implementation
and train the baseline model on the same image dehazing dataset which contains 6,900 hazy images.
We randomly choose 495 synthetic hazy images (which are not used in the training stage) for evaluations.
Table~\ref{tab: effecf-constraint} shows the experimental results of all the baseline algorithms and one example is shown in Figure~\ref{fig: effect-loss-function}.
As shown in Figure~\ref{fig: effect-loss-function}(b), the BaseGAN method does not generate clear images.
The structures of the restored image (e.g., the sky) by the BaseGAN method are significantly different from the input when the physics model is not used.
In contrast, the proposed method with the image degradation model constrained learning process generates much clearer images.
%

%
We show the effect of the proposed loss functions in Table~\ref{tab: effecf-constraint}.
We note that the method without loss functions~\eqref{eq: pixel-wise-loss-co},~\eqref{eq: pixel-wise-loss}, or~\eqref{eq: pixel-wise-loss-co-2} is less effective
and using one of these helps obtain better results.
The restored images in Figure~\ref{fig: effect-loss-function}(c)-(f) demonstrate that the proposed algorithm with the proposed loss function is able to generate clear images.

\vspace{-2mm}
{\flushleft {\bf Relation with GAN-based methods.}}
Recently, several methods have been proposed to improve the GAN~\cite{GAN}, e.g., CycleGAN~\cite{CycleGAN2017}, DiscoGAN~\cite{DiscoGAN}, and DualGAN~\cite{dualGAN}.
The CycleGAN algorithm~\cite{CycleGAN2017} uses two generators and two discriminators
for image-to-image translation when the paired training data is not available.
In addition, a cycle consistency loss is used to train the proposed CycleGAN network.
The other methods~\cite{derain/gan,pixel/to/pixel,SRGAN,gan/map,Xu_2017_ICCV} explore GANs in the conditional settings
and use a pixel-wise loss function (i.e.,~\eqref{eq: pixel-wise-loss}) with a perceptual loss function~\cite{Perceptual/Loss} to ensure that the outputs of the generative network
are close to the ground truth images in the training stage.

Although these methods are not designed for the image restoration problems addressed in this work,
we train the most related algorithms using the same training datasets as the proposed approach to clarify the differences.
As these algorithms directly learn the inverse process (i.e.,~\eqref{eq: mapping-function}) by end-to-end trainable networks without considering the image formation process, they are not able to generate clear images as demonstrated in Section~\ref{ssec: image-deblurring}-\ref{ssec: Image Deraining}.

Using the notations in the proposed method, the CycleGAN algorithm assumes that $\mathcal{F}(\mathcal{G}(y_i)) = y_i$ and $\mathcal{G}(\mathcal{F}(x_i)) = x_i$,
where $\mathcal{F}$ is a generator which has the similar effect to the mapping function $\mathcal{H}$ in the physics model~\eqref{eq: mapping-function-low-level}.
With this assumption, this algorithm is likely to converge to a trivial solution as the identity mapping functions always hold for this assumption.
This is the main reason why the results by the CycleGAN algorithm are similar to the inputs, e.g., the dehazing results are similar to the hazy inputs
(Figure~\ref{fig: dehazing-real}(g)).

Different from the CycleGAN algorithm, our method does not learn the generator $\mathcal{F}$ as the physics model is known.
Thus, our method is able to avoid trivial solutions and performs favorably against the state-of-the-art algorithms on each task.

As the proposed method uses loss functions~\eqref{eq: pixel-wise-loss-co},~\eqref{eq: pixel-wise-loss}, and~\eqref{eq: pixel-wise-loss-co-2}, it is of interest to analyze the performance gain if the same loss functions are used in the PCycleGAN method.
Using the same image dehazing datasets,
we train a model using the PCycleGAN method with the same loss functions as the proposed algorithm for evaluation.
Table~\ref{tab: effecf-constraint} and Figure~\ref{fig: effect-loss-function}(g) show that the PCycleGAN method with the same loss functions does not generate clear images.

%
\begin{table*}[!t]
\caption{\label{tab: effecf-constraint} Effectiveness of the proposed algorithm and loss functions on the image dehazing task.
}
\vspace{-3mm}
\centering
\begin{tabular}{ccccccccccc}
\toprule
      & Input & BaseGAN & w/o~\eqref{eq: pixel-wise-loss-co} &  w/o~\eqref{eq: pixel-wise-loss} & w/o~\eqref{eq: pixel-wise-loss-co-2} & w/o~\eqref{eq: pixel-wise-loss-co} \&~\eqref{eq: pixel-wise-loss-co-2} & PCycleGAN w/~\eqref{eq: pixel-wise-loss-co}-\eqref{eq: pixel-wise-loss-co-2} & Ours\\
\midrule
PSNR   & 18.52 & 29.90  & 31.93 & 21.24 & 31.71  & 30.53& 27.46 &\bf{32.05} \\
SSIM   & 0.8357& 0.9541 &0.9718 & 0.8096& 0.9693 & 0.9694&0.9408& \bf{0.9722} \\
\bottomrule
\end{tabular}
\vspace{-1mm}
\end{table*}
%

\begin{figure}[!t]\footnotesize
\centering
\begin{tabular}{cccc}
\includegraphics[width = 0.45\linewidth]{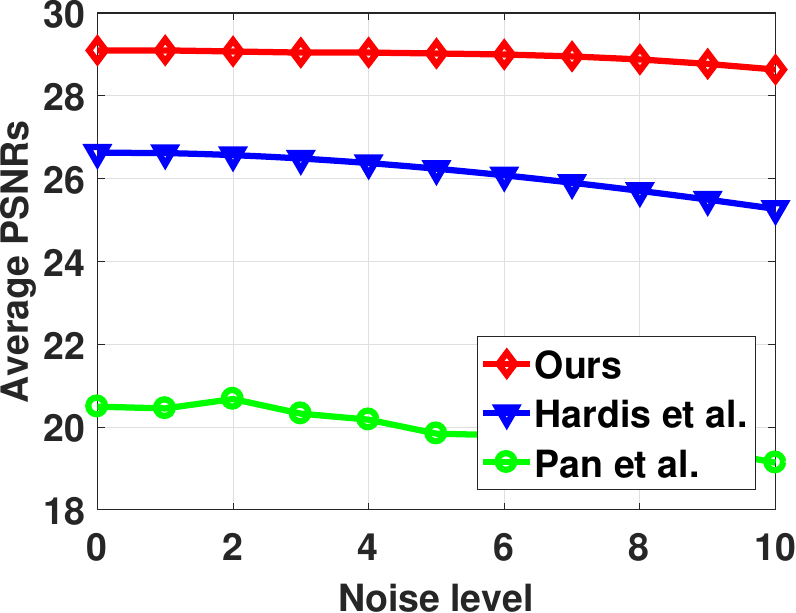} &
\includegraphics[width = 0.45\linewidth]{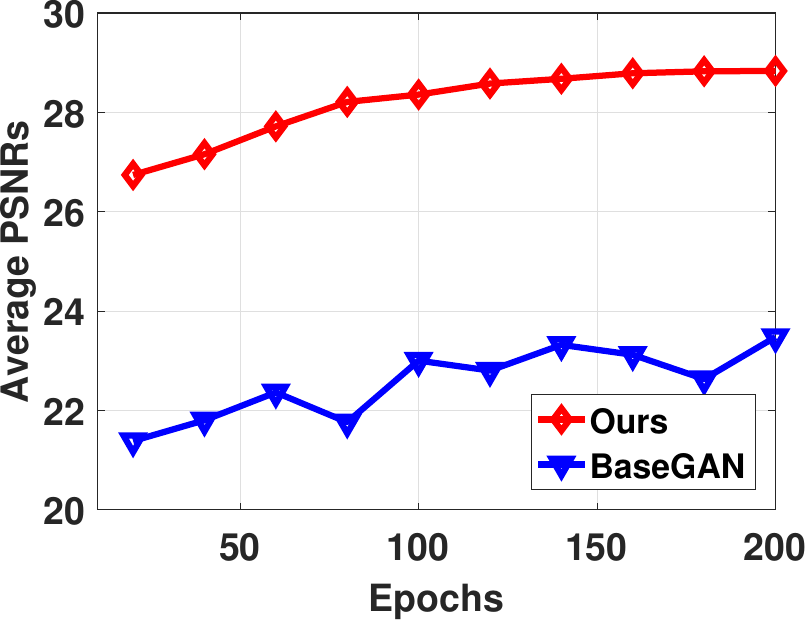}\\
(a) & (b)\\
\end{tabular}
\vspace{-3mm}
\caption{
(a) Quantitative evaluations on the blurred text images with random noise. The proposed method is robust to image noise in image deblurring.
(b) Quantitative evaluations of the convergence property on the blurred text images.
}
\label{fig: robust-to-noise}
\vspace{-2mm}
\end{figure}

\vspace{-3mm}
{\flushleft {\bf Robustness to image noise.}}
We evaluate the proposed method using 100 clear text images from~\cite{CNN/text/deblur}, where each sample is blurred and added with random noise ranging from 0\% to 10\%.
Figure~\ref{fig: robust-to-noise}(a) shows that the proposed method performs well even when the noise level is high.

\vspace{-3mm}
{\flushleft {\bf Convergence property.}}  As our algorithm needs to jointly train generative and discriminative networks, a natural question is whether our method converges well or not.
We quantitatively evaluate the convergence properties of our method on the text image deblurring dataset~\cite{CNN/text/deblur}.
Figure~\ref{fig: robust-to-noise}(b) shows that the proposed method converges well
within 200 epochs in terms of PSNR.

We note that although using multiple discriminators in GANs may increase complexity in the training
stage, our numerical results in Figure~\ref{fig: robust-to-noise}(b) show that using the physics model makes the training process more stable and leads to better convergence performance
compared to the GAN algorithm with one discriminator and generator.

\vspace{-2mm}
{\flushleft {\bf Sensitivity analysis.}}
The proposed objective function involves the weight parameter $\lambda$ which is important for restoration tasks.
We analyze the effect of this parameter using blurred face images from the proposed
test dataset by setting its value from 10 to 100 with the step size of 10.
Table~\ref{tab: parameter-study} shows that the proposed method performs well
when $\lambda$ is set to be within a wide range (i.e.,  [40, 100]) and
slightly better when this parameter is set to be $50$.

\begin{table*}[!t]
\caption{\label{tab: parameter-study} {Sensitivity analysis with respect to the weight parameter $\lambda$ and the number of ResBlocks.}}
\vspace{-3mm}
\centering
\begin{tabular}{ccccccccccc}
 \multicolumn{11}{c}{{\footnotesize Effect of weight parameter $\lambda$ on image deblurring}}           \\
\toprule
$\lambda$    &    10    &    20    &    30   &   40    &    50      &    60       &   70        &   80        &   90       &   100\\
\midrule
PSNR   &   22.92  &   23.77  &   24.06 &   24.25   &  \bf{24.57}   &    24.36    &    24.55    &   24.54     &  24.41     &   24.55\\
SSIM   &   0.6782 &  0.7103  &  0.7378 &  0.7489   & \bf{0.7531}   &    0.7528   &   0.7489    &  0.7521     &  0.7477    &   0.7488\\
\bottomrule
\end{tabular}
\begin{tabular}{ccccccccccc}
 \multicolumn{11}{c}{{\footnotesize Influence of the number of ResBlocks (\#) on image deblurring}}           \\
\toprule
 ResBlock (\#)    &     3        &    6       &    9        &    20      &   25        &    30      &    35       &   40        &   45        &   50     \\
\midrule
PSNR        &   20.45      &   20.49    &   20.77     &   20.98    &   20.33     &  20.82     &    \bf{21.22}     & 20.93      &   20.88      & 20.83    \\
SSIM        &  0.5513      &  0.5628    &   0.6005    &   0.6120   &   0.5483    &  0.6137    &    \bf{0.6173}    & 0.6062     &   0.6018     & 0.6075   \\
\bottomrule&
\end{tabular}
\vspace{-5mm}
\end{table*}

\vspace{-2mm}
{\flushleft {\bf Ablation study w.r.t. the number of ResBlocks.}}
The proposed algorithm contains several ResBlocks.
To analyze the effect of the number of ResBlocks, we evaluate the proposed network using blurred face images from the proposed test dataset by setting the number of ResBlocks from 3 to 50.
Table~\ref{tab: parameter-study} demonstrates that the proposed method is insensitive to the number of ResBlocks.
We empirically use 9 ResBlocks as a trade-off between accuracy and speed.
%

\begin{figure}[!t]
\centering
\begin{tabular}{cccc}
\includegraphics[width = 0.45\linewidth, height = 0.33\linewidth]{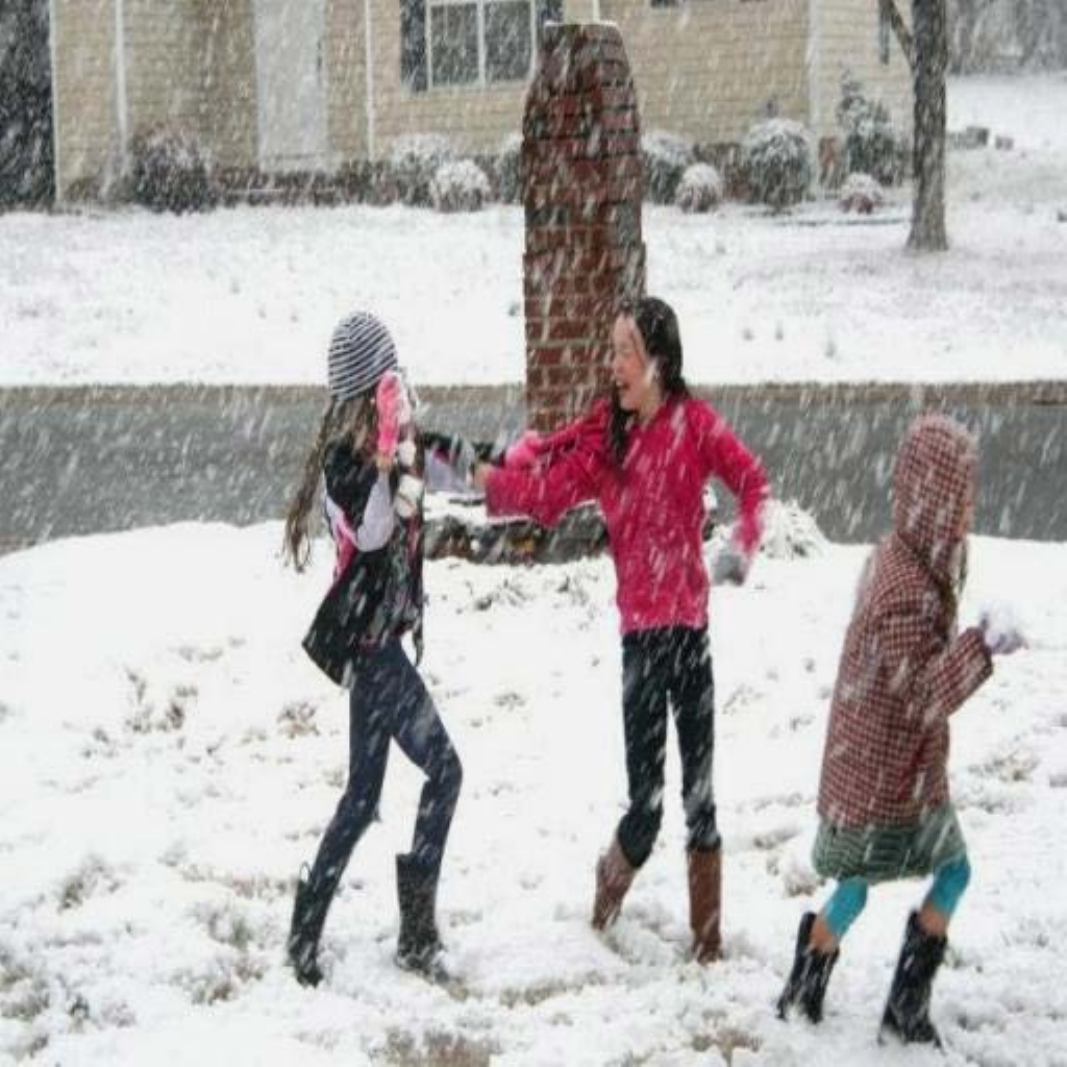} & \hspace{-4mm}
\includegraphics[width = 0.45\linewidth, height = 0.33\linewidth]{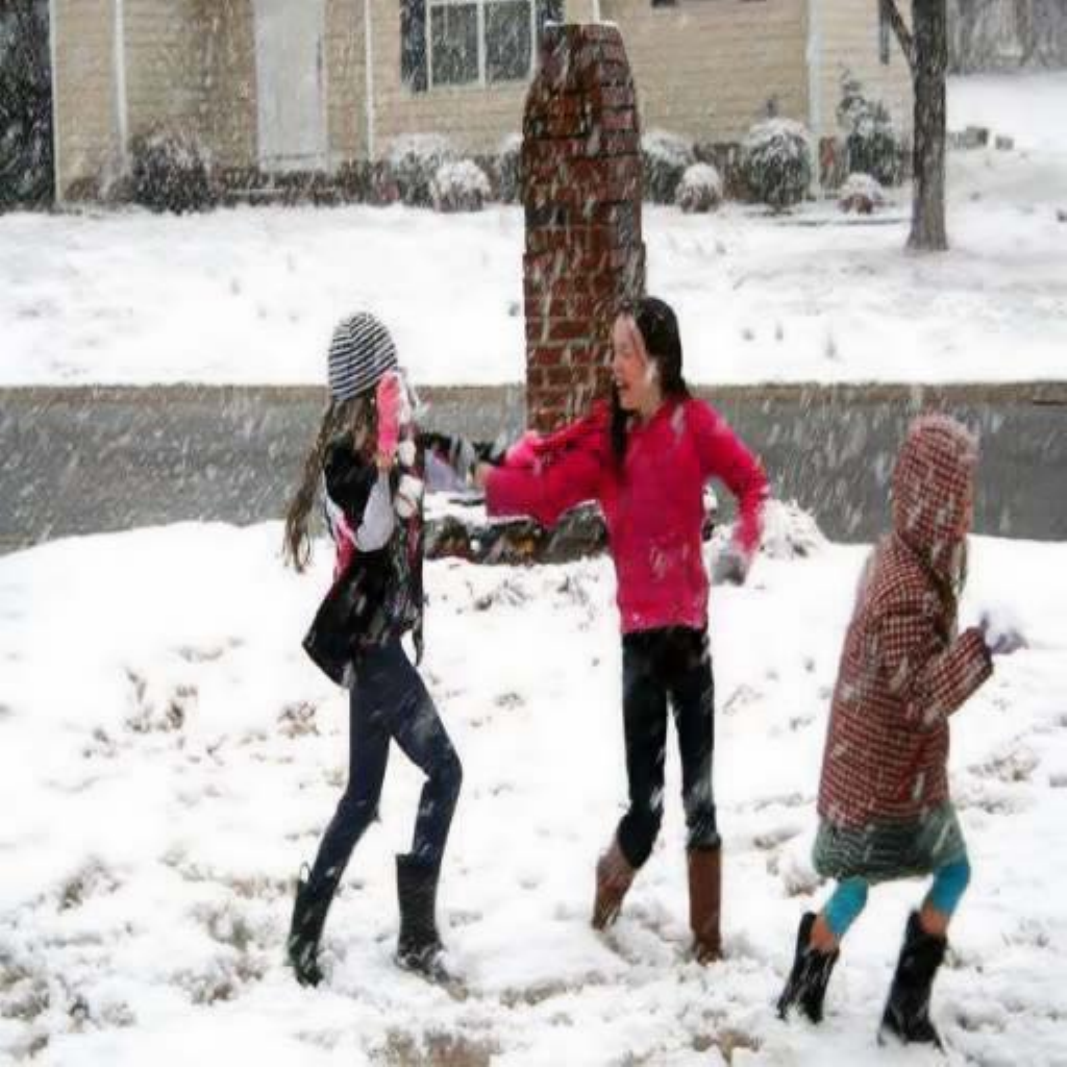}\\
(a) Input & \hspace{-4mm} (b) Restored result\\
\end{tabular}
\vspace{-1mm}
\caption{The proposed algorithm is less effective for the images
where the physics model does not describe the complex degradation process well.
}
\label{fig: limitation}
\vspace{-1mm}
\end{figure}

\vspace{-2mm}
{\flushleft {\bf Limitations.}}
Although the proposed method is able to restore images from degraded ones, it is still less effective for those examples which are caused by multiple degradation factors, e.g., both rain and haze.
For these cases, the physics model does not describe the complex image formation process well.
Figure~\ref{fig: limitation} shows an example where the proposed algorithm is not able to remove rain/snow from the input image due to the complex degradation process.
Our future work will consider jointly using semi-supervised or unsupervised learning algorithms to
address these issues.

\vspace{-3mm}
\section{Concluding Remarks}
\vspace{-1mm}
Motivated by the observation that the estimated results should be consistent with the observed inputs under the physical formation process in image restoration problems, we enforce the physics model
in a generative adversarial network for image restoration.
As the physics model is derived from the image formation process of low-level problems,
the proposed algorithm can be applied to a variety of tasks.
With an end-to-end network formulation,
the proposed algorithm performs favorably against the state-of-the-art methods
for image restoration and low-level vision problems.

\vspace{-3mm}
\section*{Acknowledgments}
\vspace{-1mm}
This work has been supported in part by the National Natural Science Foundation of China (Nos. 61922043, 61872421, 61732007),
the Natural Science Foundation of Jiangsu Province (No. BK20180471), and National Science Foundation CAREER (No. 1149783).
%
%
%
%

\vspace{-2mm}
{
\footnotesize
\bibliographystyle{IEEEtran}
\bibliography{egbib}
}

\textbf{}

%
%
%
%
%
%
%
%
\end{document}